\UseRawInputEncoding
\documentclass[11pt]{article}

\usepackage[final]{acl}

\usepackage{times}
\usepackage{latexsym}
\usepackage{enumitem}
\usepackage{array}
\usepackage[T1]{fontenc}

\usepackage[utf8]{inputenc}

\usepackage{microtype}

\usepackage{inconsolata}

\usepackage{graphicx}

\usepackage{tikz}
\usetikzlibrary{positioning,arrows.meta,shapes.geometric,fit,backgrounds,calc}
\usepackage[ruled,vlined]{algorithm2e}
\usepackage{booktabs}
\usepackage{amsmath,amssymb}
\usepackage{xcolor}
\usepackage{multirow}

\usepackage[most]{tcolorbox}
\tcbuselibrary{listings,breakable,skins}

\newtcblisting{prompt}[1][]{
  breakable,
  colback=gray!5,
  colframe=black!55,
  boxrule=0.5pt,
  arc=2pt,
  listing only,
  title=#1,
  fonttitle=\bfseries\small,
  listing options={
    basicstyle=\ttfamily\footnotesize,
    breaklines=true,
    breakatwhitespace=false,
    columns=fullflexible,
    showstringspaces=false,
  },
}

\newcommand{\methodname}{\textsc{SlideForge}}
\newcommand{\dsg}{Deck State Graph}

\title{\methodname{}: An LLM Agent for Controllable Editing of Slides as Structured Artifacts}

\author{
\textbf{Haozhen Zheng}\textsuperscript{1*},
\textbf{Fulin Wang}\textsuperscript{1*},
\textbf{Tianhu Xiong}\textsuperscript{1},
\textbf{Yingjie Yu}\textsuperscript{1},
\textbf{Shengyi Qian}\textsuperscript{2},
\\
\textbf{Hanchao Yu}\textsuperscript{2},
\textbf{Alex Schwing}\textsuperscript{1},
\textbf{Klara Nahrstedt}\textsuperscript{1},
\textbf{Mingyuan Wu}\textsuperscript{1}
\\
\textsuperscript{1}University of Illinois Urbana-Champaign,
\textsuperscript{2}Meta
\\
\texttt{\{haozhen3, fulinw2, mw34\}@illinois.edu}
}

\begin{document}
\maketitle

\begingroup
\renewcommand{\thefootnote}{\fnsymbol{footnote}}
\footnotetext[1]{Equal contribution.}
\endgroup

\begin{abstract}

Current AI agents compellingly describe slides. %
However, AI-assisted slide editing requires
more than understanding: %
the output must retain layout,
style, component structure, and native editability. Towards, AI-assisted slide editing, existing agents 
operate on screenshots or weak document representations and often fragment
coherent visual units, rasterize editable content, or break layout. %

In contrast, for controllable slide
editing we introduce an agentic framework, \methodname{}, which builds a Deck State Graph, an executable slide state that links visual decomposition, native \textsc{pptx} object structure, and perceptual organization. By recovering human-referable components while retaining fine-grained editable structure, \methodname{} supports theme-preserving reconstruction through slide-native operations and rendered-state verification. We further introduce an evaluation paradigm for controllable
slide transformation that jointly measures component recovery, preservation,
restyling consistency, visual quality, and native editability. Experiments show that \methodname{} outperforms direct prompting, screenshot-based agents, and generic code-agent baselines across
these dimensions. Code is available at \url{https://github.com/UIUC-MONET/SLIDEFORGE}.
\end{abstract}

\begin{figure}[t]
\centering
\resizebox{0.96\columnwidth}{!}{%
\begin{tikzpicture}[every node/.style={inner sep=2pt,font=\sffamily}]

\node[font=\scriptsize\bfseries\sffamily, blue!58!black, anchor=south west]
  at (0, 5.22) {(a)\enspace Slide Decomposition};

\fill[white] (0,3.2) rectangle (3,5.2);
\draw[black!35, rounded corners=2.5pt, line width=0.55pt]
     (0,3.2) rectangle (3,5.2);
\fill[blue!63!black] (0,4.86) rectangle (3,5.2);
\node[white, font=\tiny\bfseries\sffamily, anchor=west]
  at (0.1, 5.03) {Method Overview};
\fill[black!22, rounded corners=1pt] (0.12,4.55) rectangle (1.54,4.66);
\fill[black!22, rounded corners=1pt] (0.12,4.25) rectangle (1.42,4.36);
\fill[black!22, rounded corners=1pt] (0.12,3.94) rectangle (1.48,4.05);
\fill[gray!20, rounded corners=2pt] (1.73,3.30) rectangle (2.88,4.74);
\fill[blue!28, rounded corners=1pt] (1.90,4.33) rectangle (2.70,4.57);
\fill[blue!28, rounded corners=1pt] (1.90,3.88) rectangle (2.70,4.12);
\fill[blue!28, rounded corners=1pt] (1.90,3.43) rectangle (2.70,3.67);
\draw[-Stealth, gray!55, line width=0.4pt] (2.30,4.33) -- (2.30,4.12);
\draw[-Stealth, gray!55, line width=0.4pt] (2.30,3.88) -- (2.30,3.67);

\draw[-Stealth, line width=1.1pt, blue!55!black] (3.12,4.2) -- (4.18,4.2);
\node[font=\tiny\sffamily, blue!60!black, anchor=south] at (3.65,4.23)
  {Decompose};

\fill[white] (4.3,3.2) rectangle (7.3,5.2);
\draw[black!35, rounded corners=2.5pt, line width=0.55pt]
     (4.3,3.2) rectangle (7.3,5.2);
\fill[blue!63!black] (4.3,4.86) rectangle (7.3,5.2);
\node[white, font=\tiny\bfseries\sffamily, anchor=west]
  at (4.4, 5.03) {Method Overview};
\fill[black!22, rounded corners=1pt] (4.42,4.55) rectangle (5.84,4.66);
\fill[black!22, rounded corners=1pt] (4.42,4.25) rectangle (5.72,4.36);
\fill[black!22, rounded corners=1pt] (4.42,3.94) rectangle (5.78,4.05);
\fill[gray!20, rounded corners=2pt] (6.03,3.30) rectangle (7.18,4.74);
\fill[blue!28, rounded corners=1pt] (6.20,4.33) rectangle (7.00,4.57);
\fill[blue!28, rounded corners=1pt] (6.20,3.88) rectangle (7.00,4.12);
\fill[blue!28, rounded corners=1pt] (6.20,3.43) rectangle (7.00,3.67);
\draw[-Stealth, gray!55, line width=0.4pt] (6.60,4.33) -- (6.60,4.12);
\draw[-Stealth, gray!55, line width=0.4pt] (6.60,3.88) -- (6.60,3.67);

\fill[red!9]  (4.32,4.83) rectangle (7.28,5.18);
\draw[red!62!black, line width=0.75pt, rounded corners=1pt]
     (4.32,4.83) rectangle (7.28,5.18);
\node[red!65!black, font=\tiny\bfseries\sffamily, anchor=west, inner sep=1pt]
  at (4.34,5.00) {Title\;{\tiny($\tau$=1)}};

\fill[teal!9] (4.34,3.86) rectangle (5.92,4.69);
\draw[teal!72!black, line width=0.75pt, rounded corners=1pt]
     (4.34,3.86) rectangle (5.92,4.69);
\node[teal!72!black, font=\tiny\bfseries\sffamily, anchor=south west, inner sep=1pt]
  at (4.35,3.87) {Bullets\;{\tiny($\tau$=1)}};

\fill[green!7] (6.01,3.27) rectangle (7.27,4.76);
\draw[green!55!black, line width=0.75pt, rounded corners=1pt]
     (6.01,3.27) rectangle (7.27,4.76);
\node[green!55!black, font=\tiny\bfseries\sffamily, anchor=north west, inner sep=1pt]
  at (6.02,4.74) {Chart\;{\tiny($\tau$=4)}};

\draw[black!18, dashed, line width=0.45pt] (-0.1,3.02) -- (7.45,3.02);

\node[font=\scriptsize\bfseries\sffamily, orange!65!black, anchor=south west]
  at (0, 2.52) {(b)\enspace Style Adaptation};

\fill[gray!5]  (0,0.5) rectangle (3,2.5);
\draw[black!35, rounded corners=2.5pt, line width=0.55pt]
     (0,0.5) rectangle (3,2.5);
\fill[blue!63!black] (0,2.16) rectangle (3,2.5);
\node[white, font=\tiny\bfseries\sffamily, anchor=west]
  at (0.1, 2.33) {Method Overview};
\fill[black!22, rounded corners=1pt] (0.12,1.85) rectangle (1.54,1.96);
\fill[black!22, rounded corners=1pt] (0.12,1.55) rectangle (1.42,1.66);
\fill[black!22, rounded corners=1pt] (0.12,1.24) rectangle (1.48,1.35);
\fill[gray!20, rounded corners=2pt] (1.73,0.62) rectangle (2.88,2.04);
\fill[blue!28, rounded corners=1pt] (1.90,1.63) rectangle (2.70,1.87);
\fill[blue!28, rounded corners=1pt] (1.90,1.18) rectangle (2.70,1.42);
\fill[blue!28, rounded corners=1pt] (1.90,0.73) rectangle (2.70,0.97);
\draw[-Stealth, gray!55, line width=0.4pt] (2.30,1.63) -- (2.30,1.42);
\draw[-Stealth, gray!55, line width=0.4pt] (2.30,1.18) -- (2.30,0.97);
\fill[blue!63!black] (0.12,0.57) rectangle (0.34,0.50);
\fill[gray!5]        (0.36,0.57) rectangle (0.58,0.50);
\fill[black!22]      (0.60,0.57) rectangle (0.82,0.50);

\draw[-Stealth, line width=1.1pt, orange!65!black] (3.12,1.5) -- (4.18,1.5);
\node[font=\tiny\sffamily, orange!65!black, anchor=south] at (3.65,1.53)
  {Style Adapt};

\fill[gray!82!black] (4.3,0.5) rectangle (7.3,2.5);
\draw[black!35, rounded corners=2.5pt, line width=0.55pt]
     (4.3,0.5) rectangle (7.3,2.5);
\fill[orange!70!black] (4.3,2.16) rectangle (7.3,2.5);
\node[white, font=\tiny\bfseries\sffamily, anchor=west]
  at (4.4, 2.33) {Method Overview};
\fill[gray!68!white, rounded corners=1pt] (4.42,1.85) rectangle (5.84,1.96);
\fill[gray!68!white, rounded corners=1pt] (4.42,1.55) rectangle (5.72,1.66);
\fill[gray!68!white, rounded corners=1pt] (4.42,1.24) rectangle (5.78,1.35);
\fill[gray!60!black, rounded corners=2pt] (6.03,0.62) rectangle (7.18,2.04);
\fill[orange!52, rounded corners=1pt] (6.20,1.63) rectangle (7.00,1.87);
\fill[orange!52, rounded corners=1pt] (6.20,1.18) rectangle (7.00,1.42);
\fill[orange!52, rounded corners=1pt] (6.20,0.73) rectangle (7.00,0.97);
\draw[-Stealth, gray!78, line width=0.4pt] (6.60,1.63) -- (6.60,1.42);
\draw[-Stealth, gray!78, line width=0.4pt] (6.60,1.18) -- (6.60,0.97);
\fill[orange!70!black] (4.42,0.57) rectangle (4.64,0.50);
\fill[gray!82!black]   (4.66,0.57) rectangle (4.88,0.50);
\fill[gray!60!black]   (4.90,0.57) rectangle (5.12,0.50);

\draw[black!28, <->, line width=0.5pt]
     (1.50,0.40) -- node[font=\tiny\sffamily, text=black!45, anchor=north,
                          inner sep=1pt] {same content \& layout}
     (5.80,0.40);

\end{tikzpicture}%
}
\caption{%
  \textbf{\methodname{}} treats slide editing as structured presentation
  editing. \textbf{(a) Perception-aligned decomposition} constructs a
  Deck State Graph (\textsc{dsg}) by recovering perceptually coherent,
  human-referable components and grounding them in editable \textsc{pptx}
  structure. \textbf{(b) Theme-preserving reconstruction} uses the same
  \textsc{dsg} to restyle components under a target visual theme while
  preserving content, layout, and native editability.%
}
\label{fig:teaser}
\end{figure}
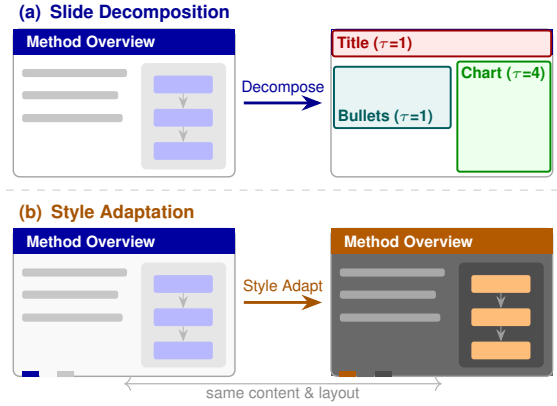

\section{Introduction}

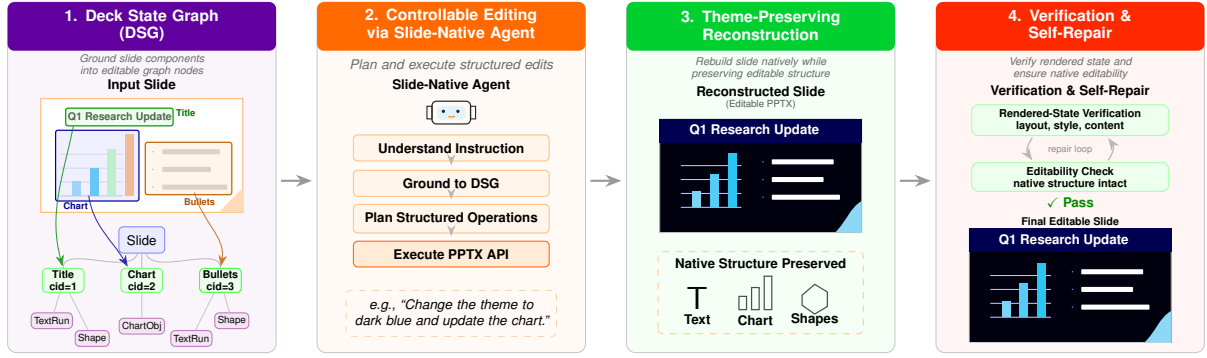
\begin{figure*}[t]
\centering
\resizebox{0.99\linewidth}{!}{%
\begin{tikzpicture}[>=Stealth,every node/.style={font=\sffamily}]

\begin{scope}[xshift=0cm]

\fill[violet!4,rounded corners=4pt] (0,0) rectangle (4.25,5.55);
\draw[black!22,rounded corners=4pt,line width=0.45pt] (0,0) rectangle (4.25,5.55);

\fill[violet!75!blue,rounded corners=3pt] (0,4.78) rectangle (4.25,5.55);
\node[white,font=\scriptsize\bfseries\sffamily,align=center,text width=3.95cm]
  at (2.125,5.18) {1. Deck State Graph\\(DSG)};

\node[font=\fontsize{4.8}{5.4}\selectfont\itshape\sffamily,
      text=black!58,align=center,text width=3.95cm]
  at (2.125,4.55) {Ground slide components\\into editable graph nodes};

\node[font=\tiny\bfseries\sffamily] at (2.125,4.23) {Input Slide};

\fill[white] (0.525,2.22) rectangle (3.725,4.02);
\draw[orange!55,line width=0.50pt] (0.525,2.22) rectangle (3.725,4.02);

\node[font=\fontsize{4.8}{5.2}\selectfont\bfseries\sffamily,
      text=black!78,anchor=west]
  at (0.82,3.72) {Q1 Research Update};

\draw[gray!42,line width=0.22pt] (0.82,2.48) -- (1.82,2.48);
\draw[gray!25,line width=0.18pt] (0.82,2.70) -- (1.82,2.70);
\draw[gray!25,line width=0.18pt] (0.82,2.92) -- (1.82,2.92);
\draw[gray!25,line width=0.18pt] (0.82,3.14) -- (1.82,3.14);

\fill[cyan!35]   (1.02,2.48) rectangle (1.16,2.71);
\fill[cyan!55]   (1.30,2.48) rectangle (1.44,2.91);
\fill[green!25]  (1.58,2.48) rectangle (1.72,3.21);
\fill[orange!50] (1.86,2.48) rectangle (2.00,3.46);

\fill[cyan!50] (2.30,3.18) circle (0.012);
\fill[cyan!50] (2.30,2.90) circle (0.012);
\fill[cyan!50] (2.30,2.62) circle (0.012);

\fill[black!18,rounded corners=0.4pt] (2.45,3.14) rectangle (3.36,3.22);
\fill[black!18,rounded corners=0.4pt] (2.45,2.86) rectangle (3.22,2.94);
\fill[black!18,rounded corners=0.4pt] (2.45,2.58) rectangle (3.48,2.66);

\fill[orange!28] (3.34,2.22) -- (3.725,2.22) -- (3.725,2.61) -- cycle;

\draw[green!60!black,fill=green!30,fill opacity=0.22,
      rounded corners=1pt,line width=0.55pt]
  (0.92,3.59) rectangle (2.62,3.87);

\draw[blue!60!black,fill=blue!30,fill opacity=0.20,
      rounded corners=1pt,line width=0.55pt]
  (0.76,2.40) rectangle (2.08,3.52);

\draw[orange!75!black,fill=orange!35,fill opacity=0.22,
      rounded corners=1pt,line width=0.55pt]
  (2.18,2.50) rectangle (3.55,3.33);

\node[font=\fontsize{4.2}{4.6}\selectfont\bfseries\sffamily,
      text=green!45!black]
  at (2.82,3.80) {Title};

\node[font=\fontsize{4.2}{4.6}\selectfont\bfseries\sffamily,
      text=blue!55!black]
  at (1.08,2.32) {Chart};

\node[font=\fontsize{4.2}{4.6}\selectfont\bfseries\sffamily,
      text=orange!65!black]
  at (3.04,2.39) {Bullets};

\begin{scope}[shift={(0,-1.43)},yscale=0.78]

\node[draw=blue!50,fill=blue!8,rounded corners=2pt,
      minimum width=0.70cm,minimum height=0.30cm,
      font=\tiny\sffamily] (Slide) at (2.125,4.12) {Slide};

\node[draw=green!60,fill=green!8,rounded corners=2pt,
      minimum width=0.66cm,minimum height=0.38cm,
      inner sep=1.0pt,
      font=\fontsize{4.8}{5.2}\selectfont\bfseries\sffamily,align=center] (Title)
  at (0.88,3.30) {Title\\cid=1};

\node[draw=green!60,fill=green!8,rounded corners=2pt,
      minimum width=0.66cm,minimum height=0.38cm,
      inner sep=1.0pt,
      font=\fontsize{4.8}{5.2}\selectfont\bfseries\sffamily,align=center] (Chart)
  at (2.125,3.30) {Chart\\cid=2};

\node[draw=green!80,fill=green!8,rounded corners=2pt,
      minimum width=0.66cm,minimum height=0.38cm,
      inner sep=1.0pt,
      font=\fontsize{4.8}{5.2}\selectfont\bfseries\sffamily,align=center] (Bullets)
  at (3.37,3.30) {Bullets\\cid=3};

\node[draw=violet!55,fill=violet!10,rounded corners=2pt,
      minimum width=0.58cm,minimum height=0.25cm,
      inner sep=1.3pt,
      font=\fontsize{4.4}{4.8}\selectfont\sffamily,align=center] (TR1)
  at (0.70,2.48) {TextRun};

\node[draw=violet!55,fill=violet!10,rounded corners=2pt,
      minimum width=0.50cm,minimum height=0.25cm,
      inner sep=1.3pt,
      font=\fontsize{4.4}{4.8}\selectfont\sffamily,align=center] (SH1)
  at (1.34,2.12) {Shape};

\node[draw=violet!55,fill=violet!10,rounded corners=2pt,
      minimum width=0.62cm,minimum height=0.25cm,
      inner sep=1.3pt,
      font=\fontsize{4.4}{4.8}\selectfont\sffamily,align=center] (CO)
  at (2.125,2.37) {ChartObj};

\node[draw=violet!55,fill=violet!10,rounded corners=2pt,
      minimum width=0.58cm,minimum height=0.25cm,
      inner sep=1.3pt,
      font=\fontsize{4.4}{4.8}\selectfont\sffamily,align=center] (TR2)
  at (2.91,2.12) {TextRun};

\node[draw=violet!55,fill=violet!10,rounded corners=2pt,
      minimum width=0.50cm,minimum height=0.25cm,
      inner sep=1.3pt,
      font=\fontsize{4.4}{4.8}\selectfont\sffamily,align=center] (SH2)
  at (3.55,2.48) {Shape};

\draw[-,gray!55,line width=0.38pt] (Slide) to[out=-105,in=90] (Title);
\draw[-,gray!55,line width=0.38pt] (Slide) -- (Chart);
\draw[-,gray!55,line width=0.38pt] (Slide) to[out=-75,in=90] (Bullets);

\draw[-,gray!45,line width=0.38pt] (Title) -- (TR1);
\draw[-,gray!45,line width=0.38pt] (Title) -- (SH1);
\draw[-,gray!45,line width=0.38pt] (Chart) -- (CO);
\draw[-,gray!45,line width=0.38pt] (Bullets) -- (TR2);
\draw[-,gray!45,line width=0.38pt] (Bullets) -- (SH2);

\end{scope}

\draw[->,green!55!black,line width=0.48pt,opacity=0.80,
      shorten >=1pt,shorten <=1pt]
  (1.05,3.62) to[out=-102,in=105] (0.88,1.33);

\draw[->,blue!60!black,line width=0.48pt,opacity=0.80,
      shorten >=1pt,shorten <=1pt]
  (1.28,2.52) to[out=-82,in=120] (1.92,1.33);

\draw[->,orange!75!black,line width=0.48pt,opacity=0.80,
      shorten >=1pt,shorten <=1pt]
  (2.95,2.57) to[out=-76,in=80] (3.37,1.33);

\end{scope}

\draw[->,line width=0.95pt,black!40] (4.33,2.72) -- (4.82,2.72);
\begin{scope}[xshift=4.90cm]

\fill[orange!4,rounded corners=4pt] (0,0) rectangle (4.25,5.55);
\draw[black!22,rounded corners=4pt,line width=0.45pt] (0,0) rectangle (4.25,5.55);

\fill[orange!82!red,rounded corners=3pt] (0,4.78) rectangle (4.25,5.55);
\node[white,font=\scriptsize\bfseries\sffamily,align=center,text width=3.95cm]
  at (2.125,5.18) {2. Controllable Editing\\via Slide-Native Agent};

\node[font=\tiny\itshape\sffamily,text=black!58,align=center,text width=3.95cm]
  at (2.125,4.56) {Plan and execute structured edits};

\node[font=\tiny\bfseries\sffamily] at (2.125,4.23) {Slide-Native Agent};

\draw[black!85,line width=0.55pt,rounded corners=2pt] (1.82,3.62) rectangle (2.43,3.98);
\fill[cyan!10,rounded corners=2pt] (1.90,3.69) rectangle (2.35,3.90);
\fill[black!85] (2.00,3.82) circle (0.026);
\fill[black!85] (2.23,3.82) circle (0.026);
\draw[orange!90!black,line width=0.42pt] (2.08,3.74) to[out=-20,in=200] (2.18,3.74);
\draw[black!85,line width=0.50pt] (1.74,3.70) rectangle (1.82,3.90);
\draw[black!85,line width=0.50pt] (2.43,3.70) rectangle (2.51,3.90);

\node[draw=orange!50,fill=orange!8,rounded corners=2pt,
      minimum width=3.10cm,minimum height=0.34cm,
      font=\tiny\bfseries\sffamily] (A1) at (2.125,3.24)
{Understand Instruction};

\node[draw=orange!50,fill=orange!8,rounded corners=2pt,
      minimum width=3.10cm,minimum height=0.34cm,
      font=\tiny\bfseries\sffamily] (A2) at (2.125,2.68)
{Ground to DSG};

\node[draw=orange!50,fill=orange!8,rounded corners=2pt,
      minimum width=3.10cm,minimum height=0.34cm,
      font=\tiny\bfseries\sffamily] (A3) at (2.125,2.12)
{Plan Structured Operations};

\node[draw=orange!75!red,fill=orange!16,rounded corners=2pt,
      minimum width=3.10cm,minimum height=0.34cm,
      font=\tiny\bfseries\sffamily] (A4) at (2.125,1.56)
{Execute PPTX API};

\draw[->,gray!55,line width=0.45pt,shorten >=2pt,shorten <=2pt]
  (A1.south) -- (A2.north);
\draw[->,gray!55,line width=0.45pt,shorten >=2pt,shorten <=2pt]
  (A2.south) -- (A3.north);
\draw[->,gray!55,line width=0.45pt,shorten >=2pt,shorten <=2pt]
  (A3.south) -- (A4.north);

\node[draw=orange!35,dashed,fill=orange!4,rounded corners=2pt,
      text width=3.05cm,align=center,
      font=\tiny\itshape\sffamily,inner sep=4pt]
  at (2.125,0.62)
{e.g., ``Change the theme to dark blue and update the chart.''};

\end{scope}

\draw[->,line width=0.95pt,black!40] (9.23,2.72) -- (9.72,2.72);

\begin{scope}[xshift=9.80cm]

\fill[green!4,rounded corners=4pt] (0,0) rectangle (4.25,5.55);
\draw[black!22,rounded corners=4pt,line width=0.45pt] (0,0) rectangle (4.25,5.55);

\fill[green!62!teal,rounded corners=3pt] (0,4.78) rectangle (4.25,5.55);
\node[white,font=\scriptsize\bfseries\sffamily,align=center,text width=3.95cm]
  at (2.125,5.18) {3. Theme-Preserving\\Reconstruction};

\node[font=\fontsize{4.8}{5.4}\selectfont\itshape\sffamily,
      text=black!58,align=center,text width=3.85cm]
  at (2.125,4.50) {Rebuild slide natively while\\preserving editable structure};

\node[font=\tiny\bfseries\sffamily,align=center] at (2.125,4.14)
{Reconstructed Slide};

\node[font=\fontsize{4.4}{4.8}\selectfont\sffamily,
      text=black!65,align=center]
  at (2.125,3.93) {(Editable PPTX)};

\fill[blue!6!black] (0.525,1.90) rectangle (3.725,3.70);
\draw[black!18,line width=0.4pt] (0.525,1.90) rectangle (3.725,3.70);

\fill[blue!32!black] (0.525,3.33) rectangle (3.725,3.70);
\node[white,font=\tiny\bfseries\sffamily,anchor=west]
  at (0.86,3.51) {Q1 Research Update};

\draw[blue!20,line width=0.20pt] (0.86,2.30) -- (1.82,2.30);
\draw[blue!20,line width=0.20pt] (0.86,2.52) -- (1.82,2.52);
\draw[blue!20,line width=0.20pt] (0.86,2.74) -- (1.82,2.74);
\draw[blue!20,line width=0.20pt] (0.86,2.96) -- (1.82,2.96);

\fill[cyan!45] (1.04,2.30) rectangle (1.18,2.55);
\fill[cyan!55] (1.32,2.30) rectangle (1.46,2.83);
\fill[cyan!65] (1.60,2.30) rectangle (1.74,3.15);

\fill[cyan!65] (2.16,3.02) circle (0.011);
\fill[cyan!65] (2.16,2.74) circle (0.011);
\fill[cyan!65] (2.16,2.46) circle (0.011);

\fill[white!78,rounded corners=0.4pt] (2.30,2.98) rectangle (3.28,3.06);
\fill[white!78,rounded corners=0.4pt] (2.30,2.70) rectangle (3.12,2.78);
\fill[white!78,rounded corners=0.4pt] (2.30,2.42) rectangle (3.38,2.50);

\fill[cyan!40] (3.30,1.90)
  .. controls (3.48,2.07) and (3.54,2.30) .. (3.725,2.40)
  -- (3.725,1.90) -- cycle;

\draw[orange!35,dashed,rounded corners=2pt,line width=0.55pt]
  (0.50,0.30) rectangle (3.75,1.60);

\node[font=\tiny\bfseries\sffamily] at (2.125,1.42)
{Native Structure Preserved};

\node[font=\Large] at (1.12,0.86) {$\mathsf{T}$};
\node[font=\tiny\bfseries\sffamily] at (1.12,0.52) {Text};

\draw[black!70,line width=0.55pt] (1.78,0.68) rectangle (1.90,0.89);
\draw[black!70,line width=0.55pt] (1.98,0.68) rectangle (2.10,1.03);
\draw[black!70,line width=0.55pt] (2.18,0.68) rectangle (2.30,1.21);
\node[font=\tiny\bfseries\sffamily] at (2.04,0.52) {Chart};

\draw[black!70,line width=0.55pt]
  (2.78,0.91) -- (2.98,1.08) -- (3.18,0.96) -- (3.18,0.73)
  -- (2.98,0.61) -- (2.78,0.75) -- cycle;
\node[font=\tiny\bfseries\sffamily] at (2.98,0.52) {Shapes};

\end{scope}

\draw[->,line width=0.95pt,black!40] (14.13,2.72) -- (14.62,2.72);

\begin{scope}[xshift=14.70cm]

\fill[red!4,rounded corners=4pt] (0,0) rectangle (4.25,5.55);
\draw[black!22,rounded corners=4pt,line width=0.45pt] (0,0) rectangle (4.25,5.55);

\fill[red!68!orange,rounded corners=3pt] (0,4.78) rectangle (4.25,5.55);
\node[white,font=\scriptsize\bfseries\sffamily,align=center,text width=3.95cm]
  at (2.125,5.18) {4. Verification \&\\Self-Repair};

\node[font=\fontsize{4.8}{5.4}\selectfont\itshape\sffamily,
      text=black!58,align=center,text width=3.85cm]
  at (2.125,4.50) {Verify rendered state and\\ensure native editability};

\node[font=\tiny\bfseries\sffamily] at (2.125,4.14)
{Verification \& Self-Repair};

\node[draw=green!35,fill=green!7,rounded corners=3pt,
      minimum width=3.12cm,minimum height=0.38cm,
      inner sep=2.0pt,
      font=\fontsize{4.9}{5.3}\selectfont\bfseries\sffamily,align=center]
  (V1) at (2.125,3.68)
  {Rendered-State Verification\\layout, style, content};

\node[draw=green!35,fill=green!7,rounded corners=3pt,
      minimum width=3.12cm,minimum height=0.38cm,
      inner sep=2.0pt,
      font=\fontsize{4.9}{5.3}\selectfont\bfseries\sffamily,align=center]
  (V2) at (2.125,2.78)
  {Editability Check\\native structure intact};

\draw[->,gray!55,line width=0.62pt,shorten >=1pt,shorten <=1pt]
  ([xshift=-0.58cm]V1.south)
  to[out=-125,in=125,looseness=1.25]
  ([xshift=-0.58cm]V2.north);

\draw[->,gray!55,line width=0.62pt,shorten >=1pt,shorten <=1pt]
  ([xshift=0.58cm]V2.north)
  to[out=55,in=-55,looseness=1.25]
  ([xshift=0.58cm]V1.south);

\node[font=\fontsize{4.2}{4.6}\selectfont\sffamily,
      text=black!45,align=center]
  at (2.125,3.22) {repair loop};

\node[font=\tiny\bfseries\sffamily,text=green!55!black]
  at (2.125,2.38) {$\checkmark$ Pass};

\node[font=\fontsize{4.8}{5.2}\selectfont\bfseries\sffamily]
  at (2.125,2.08) {Final Editable Slide};

\fill[blue!6!black] (0.525,0.16) rectangle (3.725,1.96);
\draw[black!18,line width=0.4pt] (0.525,0.16) rectangle (3.725,1.96);

\fill[blue!32!black] (0.525,1.59) rectangle (3.725,1.96);
\node[white,font=\tiny\bfseries\sffamily,anchor=west]
  at (0.86,1.77) {Q1 Research Update};

\draw[blue!20,line width=0.20pt] (0.86,0.56) -- (1.82,0.56);
\draw[blue!20,line width=0.20pt] (0.86,0.78) -- (1.82,0.78);
\draw[blue!20,line width=0.20pt] (0.86,1.00) -- (1.82,1.00);
\draw[blue!20,line width=0.20pt] (0.86,1.22) -- (1.82,1.22);

\fill[cyan!45] (1.04,0.56) rectangle (1.18,0.81);
\fill[cyan!55] (1.32,0.56) rectangle (1.46,1.09);
\fill[cyan!65] (1.60,0.56) rectangle (1.74,1.41);

\fill[cyan!65] (2.16,1.28) circle (0.011);
\fill[cyan!65] (2.16,1.00) circle (0.011);
\fill[cyan!65] (2.16,0.72) circle (0.011);

\fill[white!78,rounded corners=0.4pt] (2.30,1.24) rectangle (3.28,1.32);
\fill[white!78,rounded corners=0.4pt] (2.30,0.96) rectangle (3.12,1.04);
\fill[white!78,rounded corners=0.4pt] (2.30,0.68) rectangle (3.38,0.76);

\fill[cyan!40] (3.30,0.16)
  .. controls (3.48,0.33) and (3.54,0.56) .. (3.725,0.66)
  -- (3.725,0.16) -- cycle;

\end{scope}

\end{tikzpicture}%
}

\caption{Overview of \methodname{}: deck state graph construction from perception-aligned slide decomposition, slide-native editing, theme-preserving reconstruction, and verification with self-repair.}
\label{fig:method_overview}
\end{figure*}

Presentation slides are a central medium for communicating knowledge in science, education, business, and design \citep{ppt1,ppt2,ppt3}. Yet despite recent advances of AI agents, reliable AI-assisted slide editing remains challenging. A central reason is a mismatch between how slides are typically perceived by multimodal agents and how they are authored and edited in practice. \textbf{A slide deck is not simply a sequence of rendered images in pixel spaces}: it is a structured artifact composed of editable text, figures, shapes, charts, tables, groups, themes, z-ordering, slide masters, and cross-slide visual conventions. 
Transforming such an artifact therefore requires more than recognizing its visual content; it requires recovering human-referable components, preserving their layout and relationships, and reconstructing the deck in a form that remains editable.

Current AI agents often approach slide editing through direct prompting, screenshot-level reasoning, or generic code generation \citep{zheng2025pptagent,talktoslides}. These approaches  produce plausible edits, but they frequently break layout, alter non-target content, lose editability, or fail to maintain deck-level design consistency. The core challenge: slide editing is not only a visual generation problem. Instead, it requires structured-artifact manipulation, where visual appearance and document structure must be jointly understood and controlled.

For this, we introduce \methodname{}, an agentic framework for controllable editing of slide decks (see Figure~\ref{fig:teaser}). %
The key novelty of \methodname{}: shift slide editing from direct multimodal generation to verified structured-artifact editing. Rather than asking a frontier model to rewrite or redraw a slide in one step within pixel space, \methodname{} decomposes the deck, grounds instructions to editable state, executes constrained operations, and verifies the resulting artifact. This design supports diverse editing tasks, including local content updates, layout refinement, generated visual replacement, style adaptation, audience adaptation, and cross-slide consistency repair.
For this, \methodname{} represents a deck as an executable slide state that connects visual appearance, editable document structure, and perceptual organization, capturing human-referable components while retaining fine-grained structure for reconstruction.

Our contributions are:%
\begin{itemize}[leftmargin=*,noitemsep,topsep=0pt]
    \item We formulate controllable slide editing as recovering and
    manipulating the structured state of a presentation, rather than
    solely regenerating its rendered appearance. %
    
    \item We introduce the Deck State Graph, a  representation that connects visual slide decomposition with native PPTX object structure, %
    enabling grounded reasoning. %
    
    \item We design a slide-native agentic pipeline that uses the Deck State Graph
    as an executable interface for theme-preserving reconstruction. The pipeline
    preserves native editability and uses rendered-state feedback to verify and
    repair visual or structural inconsistencies without fine-tuning frontier
    LLM/VLM backbones.
        
    \item We propose a comprehensive evaluation paradigm for controllable slide editing and show that \methodname{} improves controllability, preservation, layout correctness, restyling quality, visual quality, and editability over direct prompting, screenshot-based agents, and generic code-agent baselines across diverse editing tasks.
\end{itemize}

\section{Related Work}
\noindent \textbf{Slide generation and editing.}
Early document-to-slide systems such as PPSGen~\citep{ppsgen},
D2S~\citep{d2s}, and DOC2PPT~\citep{fu2022doc2ppt} convert documents or
scientific papers into slide decks through content selection, retrieval,
summarization, and slide structure or layout prediction. 
Recent LLM-based and agentic systems further expand slide generation: PPTAgent~\citep{zheng2025pptagent} uses reference presentations and editing actions to generate new slides; AutoPresent~\citep{autopresent} generates editable slides from natural-language instructions through tool-augmented code generation; SlideGen~\citep{slidegen} introduces a collaborative multimodal-agent pipeline with visual-in-the-loop refinement; SlideTailor~\citep{slidetailor} conditions paper-to-slide generation on user preferences inferred from example slide pairs and templates; and DeepPresenter~\citep{deeppresenter} uses environment-grounded reflection to plan, render, and revise intermediate slide artifacts. 
In contrast, \methodname{} focuses on controllably editing existing PPTX decks while preserving non-target content, layout, style, and editability.

\noindent \textbf{Structured slide and visual artifact manipulation.}
Talk to Your Slides~\citep{talktoslides} shows that language-driven slide editing benefits from manipulating structured PowerPoint objects rather than relying solely on screenshots or screen pixels. Related poster generation and editing systems such as APEX~\citep{apex}, Paper2Poster~\citep{paper2poster}, and PosterGen~\citep{postergen} also emphasize structured design operations, layout planning, and visual evaluation or feedback. \methodname{} builds on this object-centric direction, but introduces a \dsg{} that links native PPTX objects with rendered visual components, semantic roles, spatial layout anchors, logical component groups, z-order, and deck-level style metadata. This representation enables controllable edits that preserve and restyle embedded figures and rasterized diagrams, support style migration, and maintain cross-slide visual consistency.

\noindent \textbf{Graphic decomposition, layout, and evaluation.}
Graphic decomposition methods such as DeaM~\citep{dream}, LayerD~\citep{layerd}, and Qwen-Image-Layered~\citep{qwenimagelayered} recover layered RGBA representations and metadata from raster designs,
enabling layer-level editing and reuse. Layout generation methods such as LayoutGPT~\citep{layoutgpt} and LayoutDM~\citep{layoutdm} study spatial planning for visual composition, while VASCAR~\citep{vascar} uses rendered-layout feedback and automatic metrics to iteratively refine content-aware layouts. Presentation-oriented benchmarks evaluate complementary aspects of slide systems. SlidesBench~\citep{autopresent} focuses on slide generation from natural-language instructions, TSBench~\citep{talktoslides} evaluates language-driven slide editing, SlideAudit~\citep{slideaudit} targets automated detection of slide design flaws, and DECKBench~\citep{deckbench} evaluates academic slide generation and
multi-turn slide editing with instruction-following criteria. \methodname{} differs by treating slide editing as a structured-artifact task, where success requires instruction fidelity, edit locality, content preservation, layout correctness, visual quality, and native PPTX editability.

\section{Method}
\label{sec:method}
A presentation deck is both a visual artifact and an editable structured document, requiring the agent to reason about document structure before making edits. \methodname{} addresses this challenge through a two-stage design in Figure~\ref{fig:method_overview}. First, it constructs an executable deck state representation, the \dsg{}, which links rendered visual components to native PPTX objects and structural relations in Section~\ref{sec:stage1}. Second, it uses this graph inside a slide-native agentic pipeline that performs localized planning, PPTX-level editing, rendered-state verification, and targeted repair in Section~\ref{sec:stage2}. %

\subsection{Problem Formulation}
Let $D$ denote an input presentation deck in PPTX XML and $u$ a natural-language editing instruction. 
The goal is to produce an edited deck $D'$ that satisfies $u$ while preserving non-target content, layout, visual hierarchy, and native editability. 
For style adaptation, the instruction additionally specifies a target theme, and the output must restyle the deck without changing the underlying message.

\subsection{Stage I: Deck State Graph Construction}
\label{sec:stage1}

The first stage constructs a \emph{Deck State Graph} (DSG), an executable slide state that bridges three views of a presentation: native PPTX objects, rendered visual components, and human-referable editing units. This is the key novelty of \methodname{}: instead of treating slides as screenshots or as raw document XML, the DSG connects what is visible, what is editable, and what must be preserved.

\noindent \textbf{Native and visual state extraction.}
For each slide, \methodname{} parses the PPTX XML to extract native editable objects, including text boxes, shapes, tables, charts, pictures, groups, and placeholders. For each object, we record its type, bounding box, text runs, style attributes, z-order, group membership, placeholder binding, and slide-master inheritance. This object layer provides direct handles for PPTX-level editing, but it is incomplete because many meaningful regions appear only as rasterized figures, screenshots, imported diagrams, or picture objects. To recover such regions, \methodname{} also renders each slide and applies a slide-specific segmentation pipeline. We use a fine-tuned SAM-style vision-only model~\citep{ravi2024sam2}, trained on a vocabulary of $|\Lambda|=306$ slide component types. Each detected component $e_i$ stores a mask, bounding box, opaque crop, and optional component type. These rendered components are later linked to native PPTX objects when possible, allowing the agent to distinguish directly editable elements from elements that require reconstruction.

\noindent \textbf{Cross-modal grounding and refinement.}
Slide components are heterogeneous: some are easier to identify by semantic description, while others are easier to recognize visually. \methodname{} therefore uses two complementary grounding paths. In the text-mediated path, a VLM describes visually distinct regions and an LLM maps these descriptions to labels in $\Lambda$. In the visual path, a VLM directly selects applicable labels from $\Lambda$ by inspecting the rendered slide. Their union forms the slide-conditioned vocabulary $\Lambda_{\mathbf{p}}$, which guides segmentation. Then the initial detections are refined using feedback from the rendered slide state. Validated regions are whitened to produce a cleaned image $\hat{\mathbf{p}}$, and a VLM reviewer checks whether substantive content remains uncovered, whether detections are invalid, and whether overlapping candidates represent independent components. %

\noindent \textbf{Perception-aligned component state.}
Since slide elements are naturally referenced and edited at different granularities \methodname{} audits the detections into human-referable editing units. For example, a flowchart may contain many boxes and arrows but should often be edited as one diagram, while a large visual region may contain independently editable subcomponents. The final component representation is $\Omega = \left(\mathcal{E}_{\mathrm{fine}}, \mathcal{M}, \mathcal{R}\right)$
where $\mathcal{E}_{\mathrm{fine}}$ contains fine-grained detections, $\mathcal{M}$ contains perceptual merge groups, and $\mathcal{R}$ contains substantive missed regions. This preserves both instruction-level grounding and fine-grained reconstruction ability.

\noindent \textbf{Deck State Graph.}
The final DSG is a directed graph $G=(V,E)$. The node set $V$ contains slide nodes, object nodes, and component nodes. Slide nodes store background, theme, and slide-master information; object nodes store native PPTX geometry, text, style, z-order, and grouping attributes; and component nodes store the recovered visual components in $\Omega$. The edge set $E$ encodes containment, reading order, part-of hierarchy, z-order, slide-master inheritance, recurring style patterns, and edit bindings between rendered components and native PPTX objects.

Thus, the DSG acts as the executable state for controllable slide editing. It exposes human-referable targets for instruction grounding, preserves native editable handles for PPTX-level operations, and provides structural constraints for protecting non-target content. See App.~\ref{app:decompose-pipeline-prompt} for detailed prompts.

\subsection{Stage II: Slide-Native Editing Harness}
\label{sec:stage2}
Given an instruction $u$ and the Deck State Graph $\mathcal{G}$, \methodname{} treats $\mathcal{G}$ as an executable interface between user intent, rendered appearance, and native PPTX structure. We build a slide-native agent harness beyond for additional controllability.

\noindent \textbf{Grounding instructions to editable state.}
The harness first grounds $u$ to nodes and relations in $\mathcal{G}$, producing a \emph{target scope} $\mathcal{S}_{\mathrm{edit}}$ and a \emph{preservation scope} $\mathcal{S}_{\mathrm{keep}}$. This separation is central to controllability: before any modification, the agent knows which slide elements may change and which must remain fixed. Because $\mathcal{G}$ links semantic labels, text, geometry, perceptual groups, palette roles, and native PPTX objects, the same grounding step supports content edits, layout edits, style adaptation, and raster-component replacement over structured slide state rather than pixels alone.

\noindent \textbf{Executing slide-native operations.}
An LLM planner converts the grounded instruction into a small set of PPTX-level operations, such as updating text, adjusting geometry, changing styles, editing tables, regrouping objects, or replacing non-editable raster components. Whenever possible, \methodname{} edits native objects directly, so text remains editable as text, tables and shapes remain structured, and foreground elements stay anchored unless $u$ requires movement. For rasterized or imported regions, $\mathcal{G}$ still provides semantic role, bounding box, and surrounding layout context, allowing the system to reconstruct only the affected component instead of disturbing the full slide.

\noindent \textbf{Rendered-state verification and repair.}
After execution, \methodname{} renders the modified deck and verifies the output against $u$ and $\mathcal{S}_{\mathrm{keep}}$, and refines \citep{madaan2023selfrefine,liao2025longperceptualthoughtsdistillingsystem2reasoning}. The verifier checks instruction satisfaction, preservation of protected content, and presentation errors such as overflow, broken alignment, missing elements, or incorrect layering. When verification fails, the error is mapped back to the corresponding nodes in $\mathcal{G}$ and repaired locally. %

\subsection{Stage III: Style Adaptation via Theme-Preserving Reconstruction}
\label{sec:style}

Style adaptation restyles an existing deck under a target theme $\mathcal{T}$ while preserving content, layout, and native editability. \methodname{} formulates this as \emph{theme-preserving reconstruction} over the DSG: instead of regenerating slides as flat images, it reconstructs each foreground element with the document primitive that best preserves its function.

\noindent \textbf{Structured aesthetic contract.}
Given $\mathcal{T}$, \methodname{} builds a deck-level aesthetic contract $\mathcal{C}$, including role-specific backgrounds, dominant colors, role-labeled theme colors, font choices, and contrast-aware text colors. $\mathcal{C}$ provides a shared visual vocabulary for all slides, ensuring that style adaptation is globally consistent rather than independently applied to each element.

\noindent \textbf{Modality-aware reconstruction.}
Each foreground component is assigned a reconstruction modality $\tau_i \in \{0,\ldots,5\}$: discardable ornament, editable text box, editable text panel, native table, shape composition, or styled raster asset. This modality typing preserves editability by matching each component to the most suitable PPTX primitive. Text remains editable text, tables remain native tables, diagrams are reconstructed as grouped shapes, and logos, screenshots, photos, or data-critical charts are adapted as bounded raster assets.

\noindent \textbf{Palette-constrained assembly.}
For editable objects, \methodname{} constructs a functionally consistent color mapping $f_{\mathcal{C}}$ from source colors to theme colors. Identical source colors map to identical target colors, preserving semantic differentiation such as headers versus body cells, while low-contrast text is recolored using the contrast-aware palette in $\mathcal{C}$. The final deck is assembled by placing the themed background and restoring each foreground component at its DSG anchor. Geometry and layout remain fixed unless bounded text fitting is required.

\subsection{Stage IV: Verification and Self-repair}
\label{sec:verify}
Reconstruction alone does not guarantee that the edited deck is visually correct. To make slide editing reliable, we introduce a rendered verification and self repair stage that closes the loop between native PPTX editing and rendered visual feedback.

\noindent \textbf{Component verification.}
For reconstructed editable components $\tau_i \in \{1,2,3,4\}$, \methodname{} renders the single-element PPTX and compares it against the original component crop $b_i$. The verifier checks whether the reconstructed component preserves the intended content, remains inside its bounding box, and avoids local failures such as text overflow, table corruption, or shape drift. Failed components are repaired locally by revising only their parameterization, rather than regenerating the full slide.

\noindent \textbf{Slide verification.}
After assembly, \methodname{} renders the full deck and verifies it against the target theme $\mathcal{T}$, the aesthetic contract $\mathcal{C}$, and the DSG preservation constraints. The verifier checks style consistency, content preservation, readability, layout stability, layering, and native editability. When an error is detected, it is mapped back to the corresponding DSG node or relation, enabling localized repair of color, scale, z-order, or component reconstruction.

\begin{table*}[!t]
  \centering
  \small
  \setlength{\tabcolsep}{4.2pt}
  \renewcommand{\arraystretch}{0.98}
  \begin{tabular}{lcccccccc}
  \toprule
  \textbf{Method} &
  \textbf{SSIM}$\uparrow$ &
  \textbf{LPIPS}$\downarrow$ &
  \textbf{CLIP}$\uparrow$ &
  \textbf{SCAN}$\uparrow$ &
  \textbf{Cohesion}$\uparrow$ &
  \textbf{Label Fit}$\uparrow$ &
  \textbf{Gran.}$\uparrow$ &
  \textbf{Non-Red.}$\uparrow$ \\
  \midrule
  SlideCoder & 0.743 & 0.482 & 0.726 & 0.780 & 4.54 & 2.68 & 3.18 & 3.73 \\
  Claude Code (Opus 4.7) & 0.813 & 0.341 & 0.837 & 0.858 & 4.54 & 2.82 & 3.90 & 3.89 \\
  \textbf{\methodname{} (Ours)} & \textbf{0.950} & \textbf{0.054} & \textbf{0.984} & \textbf{0.960} & \textbf{4.70} & \textbf{4.50} & \textbf{4.68} & \textbf{4.87} \\
  \bottomrule
  \end{tabular}
  \caption{Decomposition results on the out-of-domain system-conference test set.
\textbf{Gran.} denotes Granularity and \textbf{Non-Red.} denotes Non-Redundancy. \methodname{} achieves the best performance on all metrics.}
  \label{tab:decomposition_results}
\end{table*}

\begin{table*}[!t]
\centering
\scriptsize
\setlength{\tabcolsep}{4.8pt}
\renewcommand{\arraystretch}{0.92}
\makebox[\textwidth][c]{%
\begin{tabular}{lccccccccc}
\toprule
\multirow{2}{*}{\textbf{Dimension}} &
\multicolumn{3}{c}{\textbf{Industrial}} &
\multicolumn{3}{c}{\textbf{Warm}} &
\multicolumn{3}{c}{\textbf{Finance}} \\
\cmidrule(lr){2-4}\cmidrule(lr){5-7}\cmidrule(lr){8-10}
& \textbf{AP} & \textbf{GI+C} & \textbf{\methodname{} (Ours)}
& \textbf{AP} & \textbf{GI+C} & \textbf{\methodname{} (Ours)}
& \textbf{AP} & \textbf{GI+C} & \textbf{\methodname{} (Ours)} \\
\midrule
Deck-level coherence
& \textbf{7.50} & 7.03 & 7.39
& \textbf{7.52} & 6.83 & 7.14
& \textbf{6.17} & 6.03 & 6.10 \\

Slide-level coherence
& 5.20 & 5.37 & \textbf{6.00}
& 5.03 & 5.07 & \textbf{5.90}
& 4.17 & 4.52 & \textbf{5.14} \\

Information fidelity
& 6.03 & 4.37 & \textbf{6.96}
& 5.38 & 4.00 & \textbf{6.55}
& 5.38 & 4.45 & \textbf{6.59} \\

Text readability
& 6.57 & 4.07 & \textbf{6.79}
& 5.14 & 3.69 & \textbf{6.38}
& 5.83 & 3.93 & \textbf{6.66} \\

Content-decoration balance
& 5.17 & 5.53 & \textbf{6.50}
& 4.21 & 5.00 & \textbf{5.97}
& 4.00 & 4.90 & \textbf{5.66} \\

Fitness for purpose
& 4.83 & 3.67 & \textbf{5.71}
& 3.62 & 2.86 & \textbf{4.90}
& 4.24 & 3.48 & \textbf{5.34} \\

\textbf{Overall}
& 5.88 & 5.01 & \textbf{6.56}
& 5.15 & 4.57 & \textbf{6.14}
& 4.97 & 4.55 & \textbf{5.91} \\
\bottomrule
\end{tabular}%
}
\caption{Restyling evaluation across three target visual styles. Scores are
reported on a 1--10 scale. AP denotes AutoPresent, and GI+C denotes
GPT-Image-2 + Claude. \methodname{} achieves the best overall score for all
styles, while also improving information fidelity, readability,
content-decoration balance, and fitness for purpose.}
\label{tab:restyling_results}
\end{table*}

\section{Evaluation Paradigm} 
\label{sec:eval_protocol}
Controllable slide editing cannot be evaluated by visual plausibility alone. A system must recover a faithful structured state and transform it without losing content, layout, style, or editability. We evaluate \methodname{} along two axes. State recovery measures whether the system reconstructs a structured representation that covers the original content, avoids fragmented components, and assigns meaningful labels. State transformation measures whether the recovered state supports theme-preserving reconstruction and editable restyling.

\begin{figure}[!t]
\centering
\includegraphics[width=\linewidth]{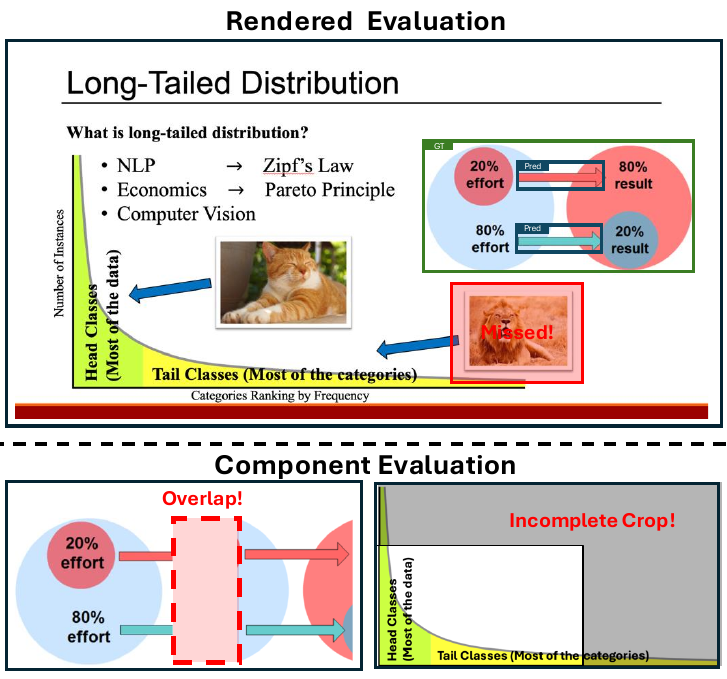}
\caption{\textbf{Complementary Decomposition Evaluation.} Rendered evaluation compares the reassembled slide with the source when ground-truth and predicted boxes differ in granularity, while component-wise evaluation detects local failures such as overlapping components and incomplete crops.}
\label{fig:decompose_eval}
\end{figure}

\subsection{State Recovery: Decomposition}
We evaluate decomposition as presentation-state recovery rather than object detection. Granularity mismatch makes box-level matching unreliable (Figure~\ref{fig:decompose_eval}), so we combine rendered evaluation for reconstructability with component-wise evaluation for overlap, crop completeness, and label validity. A valid state satisfies three properties. \emph{Reconstructability} means components recover the original artifact. \emph{Coverage} means meaningful content is captured without redundancy. \emph{Component validity} means each unit is semantically coherent and correctly typed. Our paradigm combines reference-based image similarity with reference-free VLM judgments (See App.~\ref{app:judge-prompt-decompose} for prompts), following recent practices for slide generation, decomposition, and MLLM-as-judge assessment.~\citep{tang2025slidecoderlayoutawareragenhancedhierarchical,autopresent,muppidi2025tamingllmsnegativesamples,zheng2025pptagent}

\noindent\textbf{Rendered reconstruction.}
To measure reconstructability, we rebuild a slide from the extracted components
using their predicted geometry and z-order, render the result, and compare it
with the original slide. We report MSE, PSNR, SSIM~\citep{nilsson2020understandingssim}, LPIPS~\citep{zhang2018unreasonableeffectivenessdeepfeatures}, and CLIP~\citep{radford2021learningtransferablevisualmodels} cosine
similarity, capturing pixel-level, perceptual, and semantic agreement with the
source rendering.

\noindent\textbf{Content coverage and redundancy.}
To measure whether the recovered state captures slide content without unnecessary duplication, we compute foreground coverage and spatial non-overlap. Let $F$ be the foreground mask, $b_i$ be the predicted component boxes, $B=\cup_i b_i$, and $A=\sum_i |b_i|$. We compute $R_c = |F \cap B|/|F|$ and $R_o = 1 - (A-|B|)/A$, and linearly combine them as $\mathrm{SCAN}=\lambda R_c+(1-\lambda)R_o$, where $\lambda$ is set to 0.5

\noindent\textbf{Component validity.}
We use a VLM judge to score each component crop on a 1--5 scale for
\emph{cohesion} and \emph{label fit}, measuring whether it forms a complete
editing unit and whether its assigned type matches the visual content.

\noindent\textbf{Holistic state quality.}
Finally, we evaluate global decomposition quality by showing a VLM judge the
original slide and the reconstructed rendering side by side. The judge scores
coverage, granularity, and non-redundancy on a 1--5 scale, capturing failures
that local metrics may miss, such as over-fragmented diagrams, duplicated
regions, or semantically invalid groupings.

\subsection{State Transformation: Restyling}

We evaluate restyling as the quality of a transformed presentation state, asking whether the rendered slides inhabit the target style, whether original information is preserved, and whether the output remains a genuinely editable PPTX.

\noindent\textbf{VLM aesthetic and functional judgments.}
We render each output PPTX through the same headless engine and pass all deck slides (as PNGs via LibreOffice) to a Claude-Sonnet-4.6 judge with the target style prompt. The judge scores six independent aspects on a 1 to 10 scale, each requiring a cited visual element. \emph{Deck-level coherence} rates shared visual identity across slides in palette, typography, and motif. \emph{Slide-level coherence} rates whether every element conforms to the target style. \emph{Information fidelity} rates whether chart data, labels, and diagram structure stay intact. \emph{Text readability} rates legibility of titles, body, and annotations at presentation scale. \emph{Content and decoration balance} rates whether ornament supports rather than crowds the message. \emph{Fitness for purpose} rates whether the deck would work in a realistic technical talk.

\noindent\textbf{Editability preservation.}
To measure whether the restyled artifact remains a structurally editable PPTX, we extract all leaf shapes via python-pptx, treating any non-picture shape as editable. Let $E_m$ denote the editable shape boxes of method $m$, $|b|$ the bounding-box area of $b$, and $S_w \cdot S_h$ the slide dimensions. For an output deck of $n$ slides we report
\[
R_a = \frac{\sum_{b \in E_m} |b|}{n \cdot S_w \cdot S_h}, \qquad
R_n = \frac{|E_m|}{|E_{\mathrm{GT}}|},
\]
so $R_n = 1$ for ground truth, and methods that flatten content fall well below 1.

\section{Experiments}
We follow the evaluation paradigm described in Section~\ref{sec:eval_protocol} and focus on quality of \emph{state recovery} and \emph{state transformation}.
\subsection{Dataset Collection and Pre-processing}
\noindent \textbf{Training data for SAM3.}
We collect \textbf{publicly available} presentation decks from the web, focusing on academic conference slides. This setting is challenging because academic slides contain dense text, equations, plots, tables, diagrams, and mixed raster/vector content while requiring high readability and precise layout preservation. For training the SAM3 backbone, we use 127 AI-conference decks. Bounding boxes are automatically derived from the native \textsc{pptx} structure, and each component is mapped to the SAM3 input taxonomy via an LLM-based labeling pipeline. This avoids manual box annotation and adapts to new domains by changing the taxonomy and labeling prompts.

\noindent \textbf{Evaluation data for controllable slide editing.}
To evaluate generalization, we use an out-of-domain test set of 89 system conference decks. For each deck, we include the first and last slides and uniformly sample six intermediate ones. The result remains academic in genre but differs from the AI-conference training distribution in topic, visual convention, and slide organization. Focusing on academic slides also enables a fair evaluation of style transfer. The source decks share similar communication goals, while the target styles (finance, warm, and industrial themes, with 30, 29, and 29 decks respectively) impose substantially different visual constraints.

\begin{figure*}[!t]
\centering
\includegraphics[width=0.9\linewidth]{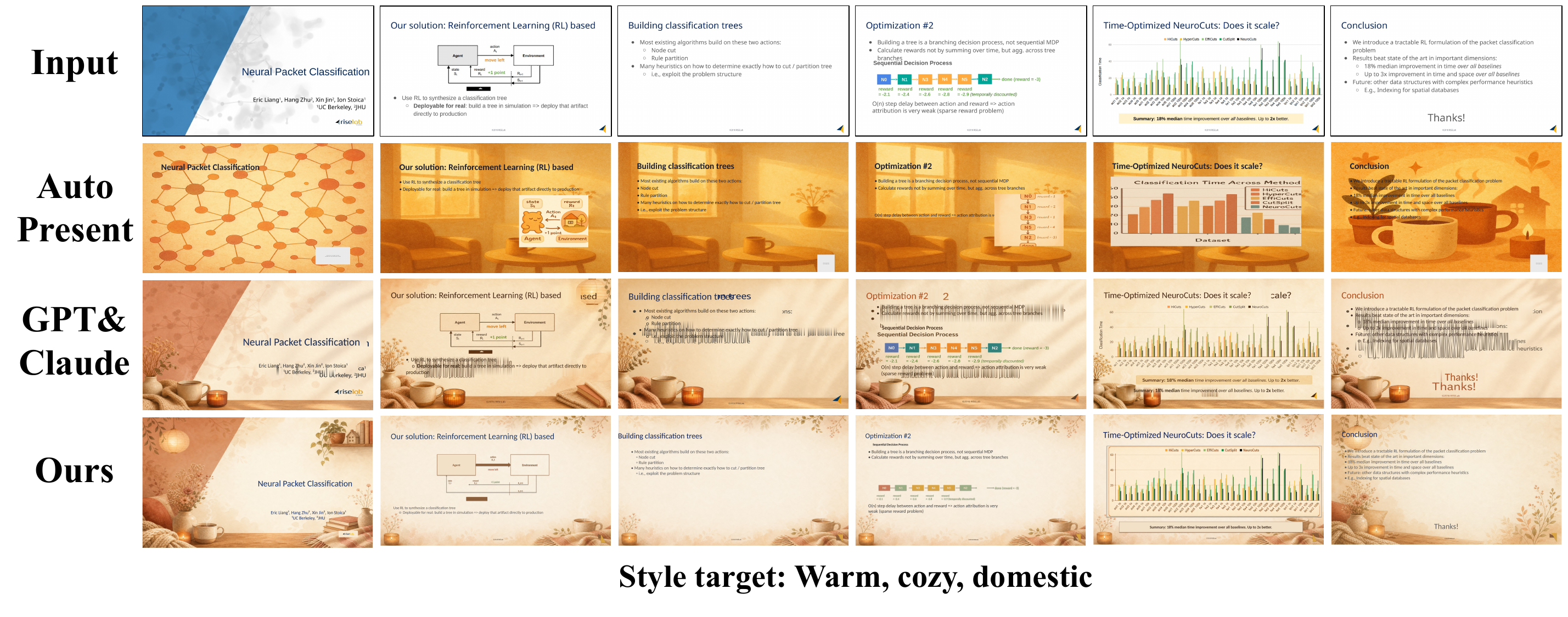}
\caption{\textbf{Case study.} AutoPresent yields distracting backgrounds and hallucinated charts; the GPT\&Claude baseline fails to fully decompose foreground elements, corrupting content and background. SlideForge preserves foreground content faithfully while cleanly restyling the background. See App.~\ref{sec:styleadpapp2} for dark industrial machinery.}
\label{caseshow}
\end{figure*}

\subsection{Baselines and Metrics}
We compare \methodname{} against prior research systems and strong
closed-source proprietary baselines, grouped by task. See App.~\ref{app:decompose-baseline-details} for detailed baseline implementations.

\begin{itemize}[leftmargin=*, nosep]
    \item \textbf{SlideCoder}~\citep{tang2025slidecoderlayoutawareragenhancedhierarchical}: a recent slide recovery pipeline used for decomposition.
    \item \textbf{Claude Opus 4.7}~\citep{anthropic2026claudeopus47}: a frontier proprietary agentic system, prompted for direct slide decomposition.
    \item \textbf{AutoPresent}~\citep{autopresent}: a prior slide-generation and editing pipeline that takes original slides image and target style prompt as input.
    \item \textbf{GPT-Image-2}~\citep{openai2026chatgptimages20} + \textbf{Claude Opus} 4.7~\citep{anthropic2026claudeopus47}: a proprietary screenshot-first pipeline that renders a full slide with GPT-Image-2, then decomposes it back into editable \textsc{pptx} with Claude Opus 4.7. It produces visually plausible slides but recovers editability only post-generation.
\end{itemize}

\noindent \textbf{Evaluation Metrics.}
We follow the evaluation protocol in Section~\ref{sec:eval_protocol}.
For decomposition, we evaluate reconstruction with \textbf{SSIM},
\textbf{LPIPS}, and \textbf{CLIP}; coverage and redundancy with \textbf{SCAN};
component validity with \textbf{Cohesion} and \textbf{Label Fit}; and holistic
state quality with \textbf{Granularity} and \textbf{Non-Red.}
For restyling, we assess source preservation, target-style adherence, and
native editability. %

\subsection{Decomposition Results}
\label{sec:decomposition_results}
Table~\ref{tab:decomposition_results} shows that \methodname{} achieves the
best performance on all decomposition metrics under the out-of-domain
system-conference test set. It improves rendered reconstruction to 0.950 SSIM,
0.984 CLIP, and 0.960 SCAN, while reducing LPIPS to 0.054, suggesting that the
recovered components can be reassembled with high visual fidelity.

Beyond reconstruction, \methodname{} also produces higher-quality editable
state: Label Fit, Granularity, and Non-Red. rise to 4.50, 4.68, and 4.87, respectively. These results indicate that DSG grounding helps recover human-referable slide units that are more coherent, accurately typed, and less
duplicative than those from generic code-agent baselines.

\subsection{Restyling Results}
\label{sec:restyling_results}
\noindent \textbf{Overall quality.} As shown in table~\ref{tab:restyling_results}, SlideForge achieves the best Overall scores of 6.56, 6.14, and 5.91 across the three style targets. By decomposing and faithfully preserving text, shapes, tables, and charts and applying generation only to the background and selected foreground elements, it leads on Slide-level coherence, Information fidelity, Text readability, and Content-decoration balance while avoiding the baselines' redundant decoration. AutoPresent's whole-slide generation yields slightly stronger Deck-level coherence but severely harms fidelity and readability and unpredictably alters key elements such as data charts. Figure~\ref{caseshow} confirms this qualitatively, where \methodname{} repaints backgrounds and accents into the target mood while keeping titles, text, charts, and diagram structure faithful, whereas the baselines flatten elements or drift from the original content. See App.~\ref{sec:styleadpapp1} for more results.

\begin{table}[ht]
\centering
\small          %
\begin{tabular}{lccc}
\toprule
\textbf{Method} & \textbf{EC} & \textbf{EAR} & \textbf{EC/input} \\
\midrule
Input        & 71.2 & 62.36\% & 1.000 \\
Autopresent  & 11.9 & 50.17\% & 0.266 \\
Gpt\_Claude  & 32.4 & 31.63\% & 0.696 \\
\textbf{ours} & \textbf{72.8} & \textbf{49.74\%} & \textbf{1.349} \\
\bottomrule
\end{tabular}
\caption{Comparison of editable count, area ratio, and count ratio against the input deck. EC is the average editable count and AR is the editable area ratio.}
\label{tab:comparison}
\end{table}

\begin{figure}[!t]
\centering
\includegraphics[width=\linewidth]{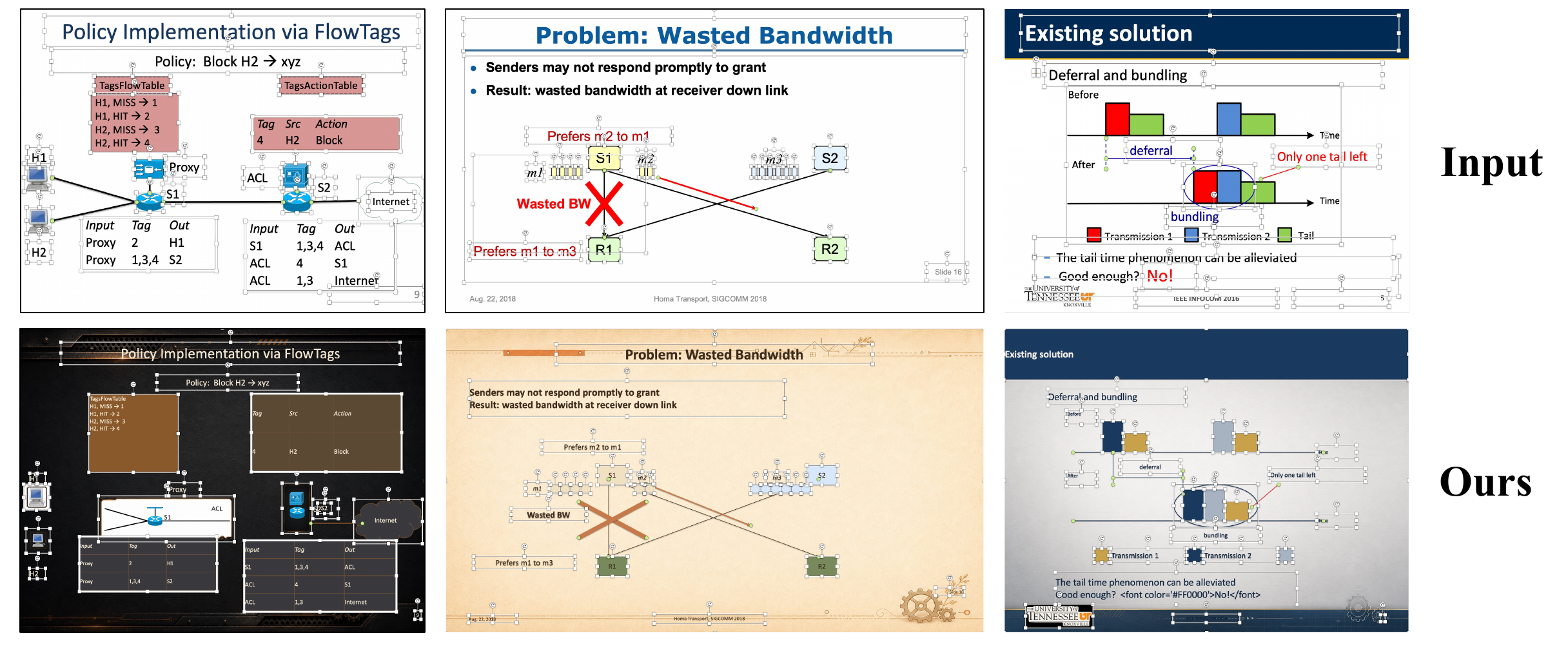}
\caption{Case study showing that the outcome PPTX of \methodname{} stays highly editable.}
\label{caseshowedit}
\end{figure}

\noindent \textbf{Editable portion.}
Table~\ref{tab:comparison} reports the editability results on the restyled decks, where we count the editable leaf shapes per slide, measure the slide area they cover, and normalize the count against the input deck. \methodname{} preserves on average 72.8 editable shapes per slide and even surpasses the input deck with a normalized editable count ratio of 1.349, whereas AutoPresent and the GPT\&Claude baseline retain only 11.9 and 32.4 shapes with editable area ratios of 0.266 and 0.696. These numbers confirm that the baselines flatten much of the deck into static imagery while \methodname{} keeps the artifact natively editable. Figure~\ref{caseshowedit} makes this concrete by opening the restyled PPTX in an editor beside the input. Text stays selectable, tables and shapes remain structured objects, and charts keep their underlying data, so a user can revise wording, recolor a shape, or update a number directly. This matches Table~\ref{tab:comparison} and shows that \methodname{} returns an editable deck rather than a flattened picture.

\noindent \textbf{Failure Analysis.}
For style adaptation, two failure modes remain (App. Figure~\ref{fig:badcase}). The $\tau=4$ shape composition modality can break visual style on overly intricate shapes, as the backbone cannot express their geometry. The $\tau=5$ raster path is limited by GPT-Image-2, which lacks transparent output, so images requiring an alpha channel fall back to a solid fill. Both failure stem from the base model's limited capacity. 

\subsection{Human Evaluation and Judge Validation}
\label{sec:human_eval}

To validate whether the automatic and VLM-based evaluations
in Sections~\ref{sec:decomposition_results} and
\ref{sec:restyling_results} reflect human judgments, we conduct a human
evaluation of the two primary tasks in our evaluation paradigm: state
recovery through decomposition and state transformation through restyling.
Eight participants took part in the study. For restyling, we stratify
30 decks from the test set, with 10 decks for each target style.
Participants evaluate the same outputs using the same six-dimensional
rubric as the VLM judge. For decomposition, participants evaluate the
same annotated component visualizations presented to the VLM judge.

\begin{table}[t]
\centering
\small
\begin{tabular}{lcc}
\toprule
Method & Restyling $\uparrow$ & Decomposition $\uparrow$ \\
\midrule
\textsc{SlideForge}             & \textbf{7.91} & \textbf{4.68} \\
AutoPresent                     & 5.34          & -- \\
GPT-Image-2 + Claude            & 6.91          & -- \\
Claude Code (Opus 4.7)          & --            & 4.09 \\
SlideCoder                      & --            & 3.07 \\
\bottomrule
\end{tabular}
\caption{Human evaluation of restyling and decomposition. Restyling is
evaluated on a 1--10 scale, while decomposition is evaluated on a 1--5
scale.}
\label{tab:human-task-eval}
\end{table}

Human judgments are consistent with the automatic and VLM-based
evaluations: \textsc{SlideForge} achieves the highest overall score for
both restyling (7.91) and decomposition (4.68), as shown in
Table~\ref{tab:human-task-eval}.

To quantify the reliability of the VLM-based evaluation, we additionally
measure agreement between judge scores and mean human scores. We report
item-level Spearman correlation, same-winner agreement, and pairwise
ranking agreement.

\begin{table}[t]
\centering
\small
\setlength{\tabcolsep}{3pt}
\begin{tabular}{lccc}
\toprule
Task & Spearman $\rho$ & \shortstack{Same\\winner} &
\shortstack{Pairwise\\ranking} \\
\midrule
Restyling      & 0.480 & 73.3\% & 72.6\% \\
Decomposition  & 0.781 & 96.0\% & 92.6\% \\
\bottomrule
\end{tabular}
\caption{Agreement between VLM-judge scores and mean human judgments.}
\label{tab:judge-human-agreement}
\end{table}

As shown in Table~\ref{tab:judge-human-agreement}, judge--human agreement
is particularly strong for decomposition, with $\rho=0.781$, 96.0\%
same-winner agreement, and 92.6\% pairwise ranking agreement. This is
consistent with decomposition criteria such as granularity, coverage, and
non-redundancy being relatively well-defined structural properties.
Restyling is more subjective, but still exhibits positive agreement across
all three measures.

Finally, we test whether the human-evaluated restyling improvements are
statistically significant. We compute per-deck score differences between
\textsc{SlideForge} and each baseline for seven metrics---the six rubric
dimensions and the overall score---across 87 evaluated decks. We apply
two-sided Wilcoxon signed-rank tests followed by Holm correction over the
14 comparisons. \textsc{SlideForge} significantly outperforms the
corresponding baseline in 13 of the 14 comparisons. Twelve comparisons
have Holm-adjusted $p<0.001$, while the comparison against
GPT-Image-2 + Claude on deck-level coherence remains significant at
$p=0.012$. The only non-significant comparison is deck-level coherence
against AutoPresent ($p=0.157$).

Beyond these task-quality judgments, we further evaluate perceptual realism
through a Turing-test-style human study in
Appendix~\ref{app:turing}.

\subsection{Controllability and Efficiency Analysis}
\label{sec:additional_analysis}

\paragraph{Element-wise controllability.}
\label{sec:controllability}
Beyond deck-level restyling quality, we evaluate whether
\textsc{SlideForge} reliably executes localized slide-native editing
operations. We evaluate every edited element from 30 sampled decks,
covering text modification, text-panel restyling, chart/table and
icon/diagram reconstruction, and generated-image replacement.
As shown in Table~\ref{tab:element-wise-editing}, text-centric operations
are particularly reliable: 91.6\% of text modifications receive a score
of at least 3, while text-panel restyling and image replacement reach
100\%. More structurally complex chart/table and icon/diagram
reconstruction achieve 72.7\% and 71.2\%, respectively. These results
show that the deck-level improvements arise from reliable localized
operations over the structured slide state rather than whole-slide
regeneration.

\begin{table}[t]
\centering
\small
\setlength{\tabcolsep}{3pt}
\begin{tabular}{lrrr}
\toprule
Operation & \# Elem. & $\geq 4$ & $\geq 3$ \\
\midrule
Text modification      & 431 & 75.2\% & 91.6\% \\
Text-panel restyling   & 24  & 95.8\% & 100\% \\
Chart/table            & 11  & 54.5\% & 72.7\% \\
Icon/diagram           & 132 & 47.7\% & 71.2\% \\
Image replacement      & 8   & 75.0\% & 100\% \\
\bottomrule
\end{tabular}
\caption{Element-wise evaluation of localized slide-native editing.}
\label{tab:element-wise-editing}
\end{table}

\paragraph{Efficiency.}
On five six-slide decks, the full pipeline costs \$4.59 per deck
and takes 1080\,s on average, including \$2.18/416\,s for
decomposition and \$2.41/664\,s for restyling. No evaluated deck
fails to converge, and restyling requires an average of only 3.0
localized self-corrections per deck; full statistics are provided
in Appendix~\ref{sec:reproducibility-runtime-cost}.

\section{Conclusion}

We introduced \methodname{}, a framework for controllable slide editing that treats decks as structured artifacts rather than rendered images. By connecting visual components, native PPTX objects, and preservation constraints through a Deck State Graph, \methodname{} enables slide-native editing, theme-preserving reconstruction, and rendered verification. Extensive experiments show that  \methodname{} improves content preservation, readability, restyling quality, and native editability.

\clearpage\newpage
\section{Limitations}
Several opportunities remain for further optimization.
The main remaining limitation in decomposition is the quality--efficiency
trade-off. \methodname{} prioritizes high-coverage, low-redundancy state recovery through iterative proposal refinement and visual review, which improves decomposition fidelity but introduces additional inference cost on content-dense academic slides.

Our pipeline also relies on frontier VLMs and rendered-state verification, which introduce substantial computational cost from both model inference and API calls. In addition, multiple verifier loops may propagate judgment errors when the verifier misidentifies visual or structural failures.

Our current task scope is also limited to controllable slide restyling and structured editing. The same framework could be extended to broader slide tasks, such as creative content rewriting, layout redesign, animation editing, and cross-deck consistency repair, but doing so would require more compute, richer tool support, and broader evaluation settings and more budgets.

\section*{Acknowledgements} This work was supported in part by NSF grants CNS-2437204, CNS-1900875, CCF-2217144, IIS-2008387, IIS-2045586, and IIS-2106825, as well as SBC UT Dallas grant 23011912 and SBC SUNY Albany grant 3-105218. This research used the Delta advanced computing and data resource, which is supported by the National Science Foundation (award OAC 2005572) 
and the State of Illinois. The opinions expressed in this publication are those of the authors and do not necessarily reflect those of the funding agencies.

\clearpage\newpage

\bibliography{custom}

\clearpage\newpage
\appendix

\section*{Appendix}

\paragraph{Appendix overview.}
The appendix provides additional evaluation, implementation details, evaluation
prompts, results, and documentation for \methodname{}. We first provide
Section~\ref{app:turing} provides an additional
\textbf{Turing-test-style human evaluation} of whether reconstructed slides exhibit perceptible machine-generated artifacts.
Section~\ref{sec:reproducibility-runtime-cost} provides
\textbf{reproducibility, runtime, and cost analysis}, including training details
for the SAM3-based component detector and end-to-end inference cost, runtime,
model-call, and retry statistics. Section~\ref{sec:inter-judge-validation}
examines \textbf{LLM-as-a-judge reliability} through independent evaluation
with Claude Sonnet 4.6 and Gemini 3.5 Flash~\citep{google2026gemini35flash}. Section~\ref{sec:element-wise-editing}
further evaluates \textbf{element-wise controllable editing}, measuring the
reliability of localized operations including text modification, chart/table
and icon/diagram reconstruction, and image replacement.

Section~\ref{sec:segpp}
details \textbf{the decomposition pipeline}, including the prompts used in Stage~I
(App.~\ref{app:decompose-pipeline-prompt}), baseline implementations for
SlideCoder and Claude Code (App.~\ref{app:decompose-slidecoder-setup};
App.~\ref{app:decompose-claude}), and the VLM-judge prompts used to evaluate
component validity and holistic decomposition quality
(App.~\ref{app:judge-prompt-decompose}). We further provide qualitative
decomposition examples in App.~\ref{app:decompose_qual}.
Section~\ref{sec:styleadpapp}
describes \textbf{the style adaptation pipeline}, including detailed quantitative
results (App.~\ref{sec:styleadpapp1}), an additional qualitative case study
(App.~\ref{sec:styleadpapp2}), limitations and failure cases
(App.~\ref{sec:styleadpapp3}), the prompts used by the style adaptation
pipeline (App.~\ref{app:style-prompts}), and the VLM-judge prompt used for
style evaluation (App.~\ref{app:style-judge-prompt}). Section~\ref{app:dataset_ethics} documents the dataset scope, privacy review,
and content-safety checks, while Section~\ref{app:llm-use} summarizes our use
of LLMs for writing support, pipeline construction, and annotation assistance. Finally, Section~\ref{app:human-eval-instructions} provides the human evaluation instructions.

\section{Turing-Test-Style Human Evaluation}
\label{app:turing}

We complement the main human evaluation in
Section~\ref{sec:human_eval} with a Turing-test-style study
that assesses whether reconstructed slides exhibit perceptible
machine-generated artifacts. Eight participants took part in
the study.

We conduct a single-stimulus Turing-test-style evaluation to measure whether
reconstructed slides exhibit perceptible machine-generated artifacts. Each
slide is presented individually, and participants are asked to classify it as
either human-made or AI-generated. Human-original slides are included to
estimate the false-alarm rate. To avoid direct pairwise comparison,
participants never observe multiple versions of the same source slide.

We report the fraction of slides judged as AI-generated, together with
$d^\prime$ and the corresponding $p$-value. Here, $d^\prime$ measures
detectability relative to the human-original false-alarm rate:
$d^\prime \approx 0$ indicates similar detectability to human originals,
whereas a larger positive value indicates that a method is more readily
recognized as AI-generated. A negative $d^\prime$ indicates that the method
is classified as AI-generated less frequently than the human originals; we
therefore conservatively interpret negative values as providing no evidence
of increased machine detectability. The $p$-value tests whether the
AI-judgment rate differs significantly from the human-original false-alarm
rate.

\begin{table}[t]
\centering
\small
\setlength{\tabcolsep}{3pt}
\begin{tabular}{lccc}
\toprule
Source & \shortstack{Judged\\as AI} & $d^\prime$ & $p$-value \\
\midrule
Human originals        & 23.6\% & --    & -- \\
\textsc{SlideForge}    & 10.7\% & -0.52 & 0.1751 \\
Claude Code            & 77.4\% & 1.47  & $<0.0001$ \\
SlideCoder             & 33.3\% & 0.29  & 0.6823 \\
\bottomrule
\end{tabular}
\caption{Turing-test-style human evaluation. ``Judged as AI'' denotes the
fraction of slides classified as AI-generated. Human originals provide the
false-alarm baseline used to compute $d^\prime$. Claude Code uses Opus 4.7.}
\label{tab:turing-human-eval}
\end{table}

As shown in Table~\ref{tab:turing-human-eval}, \textsc{SlideForge} is judged
as AI-generated in only 10.7\% of cases, compared with a 23.6\% false-alarm
rate for human originals. We do not observe a significant difference between
these rates ($p=0.1751$), providing no evidence that \textsc{SlideForge}
reconstructions are more detectable as machine-generated than human-made
slides in this setting. In contrast, Claude Code (Opus 4.7) is identified as
AI-generated substantially more often than human originals
(77.4\% vs.\ 23.6\%, $p<0.0001$). The strong separation for this baseline
indicates that the protocol is sensitive to visible machine-generated
artifacts. Together, these results suggest that \textsc{SlideForge}
substantially reduces the perceptual human-machine gap in slide
reconstruction.

\section{Reproducibility, Runtime, and Cost}
\label{sec:reproducibility-runtime-cost}

We provide additional implementation and efficiency details for
reproducibility. For the SAM3-based component detector used during slide
decomposition, we freeze the vision-language backbone and train only a
lightweight decoder adaptation. This results in 30.4M trainable parameters,
corresponding to 3.6\% of the full model. The training set contains 127
academic PPTX decks, comprising 3,510 slides, 25,132 component bounding
boxes, and 306 open-vocabulary component types. Training takes 4.7 hours on
two NVIDIA RTX 4090 GPUs. On the slide-disjoint held-out split, the detector
achieves a mean IoU of 0.873, with 95.3\% of predictions reaching
$\mathrm{IoU}@0.5$ and 88.1\% reaching $\mathrm{IoU}@0.75$.

We additionally profile end-to-end inference on five test decks, each
containing six slides. Table~\ref{tab:runtime-cost} reports average monetary
cost, wall-clock runtime, model-call count, and retry statistics per deck.

\begin{table*}[t]
\centering
\small
\resizebox{\textwidth}{!}{
\begin{tabular}{lcccc>{\raggedright\arraybackslash}p{5.0cm}}
\toprule
Stage
& Cost / deck
& Runtime / deck
& Calls / deck
& Call breakdown
& Failure / retry statistics \\
\midrule
Decomposition
& \$2.18
& 416\,s
& 74.8
& 74.8 Opus
& repair: 0; non-converged: 0 \\

Restyling
& \$2.41
& 664\,s
& 74.0
& 44.4 Opus + 24.8 Sonnet + 4.8 image
& self-correction: 3.0/deck; non-converged: 0; API retry: 0 \\

Full pipeline
& \$4.59
& 1080\,s
& 148.8
& 119.2 Opus + 24.8 Sonnet + 4.8 image
& -- \\
\bottomrule
\end{tabular}
}
\caption{Average runtime, API cost, and model-call statistics measured on
five six-slide test decks.}
\label{tab:runtime-cost}
\end{table*}

Here, a repair or self-correction is a localized modification triggered by
rendered-state verification, whereas a non-converged deck is one for which
the pipeline fails to produce a valid final artifact after the complete
retry and repair loop. In this evaluation, no deck fails to converge and no
API-level retries are required. The restyling stage invokes an average of
3.0 localized self-corrections per deck, illustrating that rendered-state
verification typically resolves errors without restarting the full
generation process.

The framework is also model-adaptive. For example, replacing Opus 4.7 with
Sonnet 4.6 for decomposition reduces the average decomposition cost from
\$2.18 to \$0.98 per deck, at the expense of a runtime trade-off. These
results make explicit the quality-efficiency trade-off of the agentic
pipeline and provide practical estimates for reproducing the full system.

\section{Inter-Judge Validation of LLM-as-a-Judge Reliability}
\label{sec:inter-judge-validation}

Because several evaluation dimensions require semantic or aesthetic
judgment, we examine whether the reported conclusions depend on the choice
of evaluator model. We independently evaluate the outputs using Claude
Sonnet 4.6 and Gemini 3.5 Flash and compare the resulting method rankings.

For restyling, the two judges produce highly consistent conclusions.
Claude Sonnet 4.6 ranks \textsc{SlideForge} first on six of the seven
reported dimensions. The only exception is deck-level coherence, where
AutoPresent receives a slightly higher score, consistent with its use of a
more uniform background treatment. Gemini 3.5 Flash ranks
\textsc{SlideForge} first on all seven dimensions.

\begin{table*}[t]
\centering
\small
\begin{tabular}{lccc|ccc}
\toprule
& \multicolumn{3}{c|}{Claude Sonnet 4.6}
& \multicolumn{3}{c}{Gemini 3.5 Flash} \\
Metric
& Ours & AutoPresent & GPT/C
& Ours & AutoPresent & GPT/C \\
\midrule
Deck-level coherence
& 6.88 & \textbf{7.06} & 6.63
& \textbf{8.51} & 7.94 & 8.16 \\

Slide-level coherence
& \textbf{5.68} & 4.80 & 4.99
& \textbf{5.95} & 3.77 & 4.86 \\

Information fidelity
& \textbf{6.70} & 5.60 & 4.27
& \textbf{6.53} & 5.12 & 2.85 \\

Text readability
& \textbf{6.61} & 5.85 & 3.90
& \textbf{6.65} & 4.49 & 2.81 \\

Content--decoration balance
& \textbf{6.04} & 4.46 & 5.14
& \textbf{6.73} & 3.28 & 3.86 \\

Fitness for purpose
& \textbf{5.32} & 4.23 & 3.34
& \textbf{5.41} & 2.83 & 1.98 \\

Overall
& \textbf{6.20} & 5.33 & 4.71
& \textbf{6.63} & 4.57 & 4.09 \\
\bottomrule
\end{tabular}
\caption{Inter-judge validation for restyling. GPT/C denotes the
GPT-Image-2 + Claude baseline. Bold indicates the best score under each
judge.}
\label{tab:interjudge-restyling}
\end{table*}

We observe the same pattern for decomposition. As shown in
Table~\ref{tab:interjudge-decomposition}, \textsc{SlideForge} obtains the
highest average score under both evaluators. The only dimension-level
exception is Claude Sonnet 4.6's cohesion score, for which Claude Code is
marginally higher. All other decomposition dimensions favor
\textsc{SlideForge} under both judges.

\begin{table*}[t]
\centering
\small
\begin{tabular}{lccc|ccc}
\toprule
& \multicolumn{3}{c|}{Claude Sonnet 4.6}
& \multicolumn{3}{c}{Gemini 3.5 Flash} \\
Metric
& Ours & Claude Code & SlideCoder
& Ours & Claude Code & SlideCoder \\
\midrule
Cohesion
& 4.552 & \textbf{4.587} & 4.463
& \textbf{4.70} & 4.54 & 4.54 \\

Label fit
& \textbf{4.343} & 2.462 & 2.288
& \textbf{4.50} & 2.82 & 2.68 \\

Granularity
& \textbf{4.620} & 3.592 & 3.364
& \textbf{4.68} & 3.90 & 3.18 \\

Non-redundancy
& \textbf{4.789} & 4.163 & 3.864
& \textbf{4.87} & 3.89 & 3.73 \\

Average
& \textbf{4.576} & 3.701 & 3.501
& \textbf{4.688} & 3.788 & 3.533 \\
\bottomrule
\end{tabular}
\caption{Inter-judge validation for decomposition. Bold indicates the best
score under each judge.}
\label{tab:interjudge-decomposition}
\end{table*}

Overall, the relative ranking of methods is stable across the two evaluator
models. The improvements of \textsc{SlideForge} therefore do not appear to
be an artifact of a particular LLM judge.

\section{Details of Element-Wise Controllable Editing Evaluation}
\label{sec:element-wise-editing}

Section~\ref{sec:controllability} reports the main results of our element-wise controllable editing evaluation. Here, we provide additional
details on the evaluation protocol.

We evaluate every edited element from 30 sampled decks. For each element,
Claude Sonnet 4.6 evaluates the final edited state on a 1--5 scale while
conditioning on the full rendered slide, the element localization, and a
local crop. We report the mean score as well as the percentages of elements
receiving scores of at least 4 and at least 3.

\begin{table*}[t]
\centering
\small
\begin{tabular}{lrrrr}
\toprule
Operation type & \# Elements & Avg. $\uparrow$ & $\geq 4$ & $\geq 3$ \\
\midrule
Text modification / recoloring / reformatting
& 431 & 3.90 & 75.2\% & 91.6\% \\

Text-label restyling with background
& 24 & 4.54 & 95.8\% & 100\% \\

Chart / table reconstruction
& 11 & 3.64 & 54.5\% & 72.7\% \\

Icon / diagram reconstruction
& 132 & 3.27 & 47.7\% & 71.2\% \\

Image update / generated-image replacement
& 8 & 4.12 & 75.0\% & 100\% \\
\bottomrule
\end{tabular}
\caption{Element-wise evaluation of slide-native controllable editing on
30 sampled decks.}
\label{tab:element-wise-editing-main}
\end{table*}

Table~\ref{tab:element-wise-editing-main} shows that text-centric operations and
image replacement are particularly reliable. Text modification achieves a
score of at least 3 for 91.6\% of 431 elements, while text-label restyling
and image replacement reach 100\% at the same threshold. Chart/table and
icon/diagram reconstruction remain more challenging, with 72.7\% and
71.2\% of elements, respectively, receiving scores of at least 3. These
failures primarily arise when complex visual elements require more precise
palette adaptation or reconstruction than the underlying models can
reliably provide. The results nevertheless show that the deck-level gains
of \textsc{SlideForge} are supported by successful localized edits across a
broad range of element types.

\clearpage
\raggedbottom

\section{Decomposition Pipeline Details}
\label{sec:segpp}
\subsection{Prompt Used in Decomposition Pipeline}
\label{app:decompose-pipeline-prompt}
\paragraph{Prompt organization.}
This section lists the prompts used in the perception-aligned decomposition stage of \methodname{}.
These prompts correspond to the Stage~I pipeline described in Section~\ref{sec:stage1}.

\begin{prompt}[Prompt 1: Text-Mediated Component Proposal]
System:
You are a vision assistant that decomposes presentation slides into their human-perceivable components. Return a list of short noun-phrase labels (2-5 words) that a human would consider one discrete visual unit.

User:
Look at this presentation slide. List every distinct component a human would treat as one whole when authoring the slide.

Granularity rules:
- A bullet list is ONE component, not one per bullet.
- A chart (including axes, title, legend) is ONE component.
- A logo + company name together may be ONE component if visually grouped.
- Decorative background shapes count as components if they are visually intentional.

Return STRICT JSON only, no prose, no code fences, with this exact shape:
{"components": ["<short phrase 1>", "<short phrase 2>", ...]}
\end{prompt}

\begin{prompt}[Prompt 2: Taxonomy Mapping for Component Proposals]
System:
You map free-form visual component descriptions onto a fixed taxonomy. You never invent new classes. If nothing fits, omit the phrase.

User:
Here is the fixed class taxonomy (use ONLY these):
{class_list_json}

Here are the phrases to map:
{phrases_json}

For each phrase, pick the single best-fit class from the taxonomy, or null if none fit reasonably.
Return STRICT JSON only, no prose, no code fences:
{{"mappings": [{{"phrase": "<phrase>", "class": "<one of the taxonomy or null>"}}, ...]}}
\end{prompt}
\begin{prompt}[Prompt 3: Visual Taxonomy Selection]
System:
You are a vision assistant. Given a slide image and a fixed class taxonomy, select every class that is actually present in the image. Be precise and prefer recall over precision, but do not include classes that are not visible.

User:
Class taxonomy (pick ONLY from these exact strings):
{class_list_json}

Look at the slide image and select every class whose instance is visibly present.
Return STRICT JSON only, no prose, no code fences:
{{"classes": ["<class 1>", "<class 2>", ...]}}
\end{prompt}
\clearpage\newpage
\begin{prompt}[Prompt 4: Rendered Coverage Review]
System:
You are a QA reviewer. You look at a 'cleaned' slide image from which an automated pipeline has tried to remove every distinct component (text, charts, photos, shapes, etc. are replaced with white). Alongside the cleaned image you may also receive a coverage overlay: the ORIGINAL slide with red rectangles drawn around every region the pipeline ALREADY extracted. You decide whether the cleaning is complete, or whether another pass of the pipeline is warranted because obvious slide components OUTSIDE the red rectangles are still visible.

User:
You are given two images:

- IMAGE 1 ("cleaned image"): the output of a component-removal pipeline. Ideally every substantive visual component (title, bullets, paragraphs, charts with their content, photographs, diagrams, captions, etc.) has been replaced with white/blank pixels.
- IMAGE 2 ("bbox overlay", may be omitted): the ORIGINAL slide with red rectangles drawn around every region the pipeline already extracted as a component. Anything inside a red rectangle has been CLAIMED as extracted.

Your job: decide whether the cleanup is acceptable, or whether it is worth running the pipeline again on this image to recover MISSED or FAILED-EXTRACTION components.

Critical rule about the red rectangles (only when IMAGE 2 is provided):
- Red rectangles show attempted extractions. If you see only faint mask-cleanup residue **inside** a red rectangle -- broken texture, dotted ghosts, scattered pixels, empty frames, thin lines, anti-aliasing halos -- DO NOT trigger a rerun for that alone.
- **Important exception**: If a substantive, readable component remains largely intact **inside** a red rectangle (e.g., a paragraph you can read, a chart with its data still visible, a recognizable photograph/diagram/table, a logo/banner with legible text), treat this as a FAILED EXTRACTION. It DOES warrant a rerun, even though it is boxed.
- For regions **outside** all red rectangles, evaluate normally, but be sensitive to importance: even a small element counts if it is clearly intentional and meaningful (a short title, a key number, a single icon, a 1-3 word label, one bullet).

Set needs_rerun = TRUE when at least one SUBSTANTIVE component is still visibly intact, either:
- (A) mostly OUTSIDE every red rectangle, OR
- (B) mostly INSIDE a red rectangle but clearly not extracted (failed extraction).

SUBSTANTIVE components (trigger rerun):
- Readable text forming a meaningful unit -- title, heading, bullet, label, caption, paragraph (even 1-3 words if clearly intentional, not random noise).
- A chart/graph with data curves, bars, fills, or legible axis labels.
- A photograph, illustration, diagram, table, infographic with recognizable content.
- A large icon, logo, or banner with readable elements.
- A key number, percentage, code, or symbol that conveys information.

Set needs_rerun = FALSE (i.e., "clean enough") when the remaining content is ONLY the following kinds of residue, regardless of location:
- Faint mask-cleanup residue inside red rectangles (as defined above).
- Chart or panel FRAMES / borders / rounded outlines with empty interiors.
- Thin horizontal or vertical LINES (axes without labels, rules, dividers, underlines).
- Random 1-3 character fragments that do not form a word/number/label.
- Thin DECORATIVE STRIPS / ribbons / bands without text.
- Anti-aliasing halos, stray pixels, tick marks, bullet dots.
- Page margins, blank background.

Positive examples (needs_rerun = FALSE / clean enough):
- Empty chart frames with no curves/labels, plus a thin colored bar at bottom edge, plus a 2-letter random fragment -> FALSE.
- Blank slide with just a thin footer line -> FALSE.
- Only axis lines and tick marks visible -> FALSE.
- A red-boxed photograph area shows only broken texture/ghost, no recognizable subject -> FALSE (already extracted).

Negative examples (needs_rerun = TRUE):
- A full chart with its curve and labels still visible, outside any red rectangle -> TRUE.
- A paragraph or bullet text with most words intact, outside any red rectangle -> TRUE.
- A photograph of an object clearly visible, outside any red rectangle -> TRUE.
- A red-boxed region still contains a readable paragraph (extraction failed) -> TRUE.
- A single-word title "Results" outside any box, clearly intentional -> TRUE.
- A small key number "42%

Return STRICT JSON only, no prose, no code fences:
{"needs_rerun": true|false, "reason": "<one short sentence>", "remaining_components": ["<short phrase>", ...]}
\end{prompt}
\clearpage\newpage
\begin{prompt}[Prompt 5: Component Validity and Text Extraction]
System:
You review an extracted slide-component crop. You decide: (1) whether the crop contains meaningful content or is essentially blank / noise / residue; (2) whether the crop is TEXT-ONLY (the entire component is just text, with no surrounding diagram / shape / chart / icon / photo context); and (3) when text-only, you transcribe the text into LaTeX (preserving any formulas, sub/superscripts, symbols). This crop is a bbox rectangle from the original slide -- it is NOT mask-cut -- so the pixels show the actual rendered region, possibly with some surrounding whitespace from the slide.

User:
You are given:

- IMAGE 1: ONE extracted component crop from a slide-decomposition pipeline.
- IMAGE 2 (optional, when present): the FULL ORIGINAL SLIDE with a red
  rectangle outline showing exactly where this crop's bbox sits on the
  slide. Use it to disambiguate narrow / sparse / character-shaped crops
  (e.g. brackets, mathematical operators, single letters, dividers,
  vertical bars) where IMAGE 1 alone is hard to interpret. The slide
  context tells you what content the bbox actually encloses and what
  surrounds it.

The pipeline believed IMAGE 1 held an instance of class "{class_name}" (filename: {filename}).
IMAGE 1 is an opaque rectangular region from the slide (bbox-only, no masking).

When IMAGE 2 is provided, ALWAYS look at it before judging IMAGE 1 --
a thin or partial-looking crop is often a fully legitimate character /
delimiter / divider whose role is only clear in context.

Answer FOUR things:

1. valid (bool): Is this crop a real component?
   - FALSE when the crop is essentially blank, only 1-3 stray character fragments, a faint smeared ghost, a thin line, anti-aliasing halo, or pure edge debris.
   - TRUE when you can see at least one recognisable slide element: readable text (even a single word), a shape, icon, logo, arrow, chart fragment, diagram, photograph, or a coloured region that is clearly a deliberate graphic.

2. is_text_only (bool): Is this component composed ENTIRELY of text?
   - TRUE when the component is just rendered text (a title, a label, a paragraph, a caption, a formula, a numeric value, etc.) with no graphical element beyond the typographic shapes themselves.
   - FALSE when the component contains non-text graphics as part of the element: e.g. an icon with a label, a chart with axis numbers, a diagram with callouts, a labeled shape, a photograph with watermark text. The presence of any non-text graphical content that is part of the component makes it NOT text-only.
   - Only set is_text_only = TRUE if valid is also TRUE. If the crop is invalid (blank / residue), set is_text_only = FALSE.

3. ocr_latex (string): If (and only if) is_text_only is TRUE, transcribe the visible text into LaTeX source.
   - Preserve formulas: e.g. $E = mc^2$, \frac{{a}}{{b}}, x^{{2}}, \sum_{{i=1}}^{{n}} a_i.
   - Preserve line breaks with \\.
   - Preserve bullet markers as \item or plain hyphens inside an itemize-style block if you see a list.
   - Use plain LaTeX escaping for special characters ( %
   - If is_text_only is FALSE, return an empty string for ocr_latex.

4. bbox_is_tight (bool): Does the bbox tightly enclose the actual content, or does it
   include a noticeable amount of empty whitespace beyond the content?
   - This question matters ONLY when is_text_only is TRUE. For non-text crops, set
     bbox_is_tight = TRUE (the pipeline does not rewrite non-text bboxes).
   - For text crops:
       - TRUE: the text fills most of the crop's width AND height; trailing /
         leading / top / bottom whitespace is small (each side roughly less
         than ~10%
       - FALSE: there is substantial whitespace on one or more sides -- the
         text occupies clearly less than the crop's full extent (e.g. the text
         ends well before the right edge and a large white strip remains, or
         the text sits in the upper portion with a tall blank band below).
         A typical sign: the bbox extends ~30%
         one axis and the empty area could fit other slide content.
   - Set bbox_is_tight = TRUE if uncertain or marginal -- only flag FALSE when
     the whitespace is obvious and substantial.

5. bbox_is_too_small (bool): Is the bbox too small, causing the text content to be
   visibly cut off or clipped at the edges?
   - This question matters ONLY when is_text_only is TRUE. For non-text crops, set
     bbox_is_too_small = FALSE.
   - For text crops:
       - TRUE: the bounding box is too restrictive. Text glyphs, descenders,
         ascenders, or characters at the very edges are visibly chopped off.
       - FALSE: the text is fully contained within the crop. No parts of the
         characters are clipped by the edges.
   - Set bbox_is_too_small = FALSE if uncertain -- only flag TRUE when the text is
     clearly missing parts due to the crop boundaries. Note: A bounding box
     cannot be both "too small" (TRUE here) and have substantial whitespace
     (FALSE for bbox_is_tight).

Return STRICT JSON only, no prose, no code fences:
{{"valid": true|false, "is_text_only": true|false, "ocr_latex": "<latex string or empty>", "bbox_is_tight": true|false, "bbox_is_too_small": true|false, "reason": "<one short sentence>"}}
\end{prompt}

\begin{prompt}[Prompt 6: Missed Region Validation]
System:
You judge whether a small visual element inside a slide is a STANDALONE component or just a SUB-PART of a larger surrounding element. An icon or shape drawn inside a larger diagram is a sub-part of the diagram, not a standalone component. A title sitting above a chart is standalone.

User:
Look at the slide image. Two overlapping rectangular regions were detected by a grounding model:

- Region A (smaller, class="{class_a}"): bbox xyxy = {bbox_a}
- Region B (larger, class="{class_b}"): bbox xyxy = {bbox_b}

Region A is contained (fully or mostly) inside Region B.

Decide:

- is_part_of = TRUE if the content inside Region A is a sub-element, a decoration, or an integral visual piece of Region B's content. That is, treating B as a whole already visually includes A; separating A out would damage B's meaning. Examples: A is an icon drawn inside a diagram; A is a small shape that's part of a bigger illustration; A is a caption/label that belongs with a callout body.

- is_part_of = FALSE if Region A is an independently-authored component that merely happens to spatially overlap B. Examples: A is a standalone title, a caption or paragraph sitting on top of a decorative backdrop; A is a separate bullet/icon placed near B; A is a textual label sitting next to but not "inside" B's content.

Return STRICT JSON only, no prose, no code fences:
{{"is_part_of": true|false, "reason": "<one short sentence>"}}
\end{prompt}

\clearpage\newpage
\begin{prompt}[Prompt 7: Missed Region Localization]
System:
You review a slide-decomposition pipeline's coverage. The pipeline occasionally misses regions outright -- substantive content sits outside every extracted bbox. Your sole job in this call is to identify those missed regions so they can be added as first-class components.

User:
You are given:

- IMAGE 1: the original slide.
- IMAGE 2: the same slide with a red rectangle drawn over every bbox the
  pipeline currently extracted (combined across all iterations, after
  cross-iter dedup and per-component validity pruning).

For reference, here is the structured list of those existing bboxes:

{components_json}

Allowed taxonomy classes (use ONLY these exact strings for `class`):
{taxonomy_json}

Identify SUBSTANTIVE content visible OUTSIDE every red rectangle that
should have been a component (e.g. a callout label, a sub-diagram block,
an output text label connected by a dashed arrow, a body paragraph the
pipeline did not detect, a "Predictions" / "Output" terminal box at the
end of a flowchart).

For each missed region provide:

- "bbox": [x1, y1, x2, y2] in ORIGINAL slide pixel coordinates.
- "class": one taxonomy class (MUST be from the list above).
- "description": one short sentence describing the missed content.

CRITICAL RULE -- report regardless of downstream merging:

Always report a missed region whenever a HUMAN annotator would call
it its own component, EVEN IF you suspect it will later be merged
into a larger semantic group (e.g. you think the "Output" callout at
the end of a flowchart will be merged into the flowchart in the next
step). The pipeline keeps both granularities: every missed region is
saved as its own fine-grained component crop on disk, AND the next
step may additionally include it in a merge_group. Skipping a missed
region here permanently loses its standalone fine crop. Lean toward
OVER-reporting in this phase; the next phase deduplicates by merging.

DO report:
- Standalone text labels / paragraphs not in any red rectangle.
- Output / terminal callouts of a flowchart connected by an arrow --
  even if you think they semantically belong to the flowchart, report
  the bbox of just the callout text + its arrow here.
- Sub-elements of a diagram that ended up uncovered -- report each
  uncovered sub-element separately, not as one giant union bbox.
- Any content a human would consider its own component.

DO NOT report:
- Anti-aliasing fringes, mask-cleanup residue, or stray pixels inside
  existing red rectangles.
- Empty page margins, decorative strips along edges.
- Orphan 1-3 character letter fragments.
- Frames / outlines with empty interiors.
- Thin connector arrows / lines on their own (only report a callout
  arrow if it is part of a textual / labelled callout).

Note: do NOT make merge decisions in this call. Just list missed
regions at fine granularity. A separate later step will decide whether
the missed regions plus existing components should be combined into
larger semantic units. Trust the two-step design: report fine here,
merge later.

Return STRICT JSON only, no prose, no code fences:
{{
  "missed_regions": [
    {{"bbox": [x1,y1,x2,y2], "class": "...", "description": "..."}}
  ]
}}

If nothing is missed, return ``{{"missed_regions": []}}``.
\end{prompt}

\clearpage\newpage
\begin{prompt}[Prompt 8: Point-Based Missed Region Grounding]
System:
You review a slide-decomposition pipeline's coverage. Substantive content sometimes sits outside every extracted bbox. Your job in this call is to locate those MISSED regions by giving ONE anchor POINT per missed region -- SAM (a downstream segmentation model) will use that point to ground the actual mask + bbox. You DO NOT estimate bboxes yourself (VLMs are unreliable at pixel-precise bbox coordinates); you only point at where the missed content sits on the slide, and SAM does the geometric grounding.

User:
You are given:

- IMAGE 1: the original slide.
- IMAGE 2: the same slide with a red rectangle drawn over every bbox the
  pipeline currently extracted (combined across all iterations, after
  cross-iter dedup and per-component validity pruning).

For reference, here is the structured list of those existing bboxes:

{components_json}

Allowed taxonomy classes (use ONLY these exact strings for `class`):
{taxonomy_json}

Identify SUBSTANTIVE content visible OUTSIDE every red rectangle that
should have been a component (e.g. a callout label, a sub-diagram block,
a "Predictions" / "Output" terminal box at the end of a flowchart, a
body paragraph the pipeline did not detect, a separate photograph).

For each missed region, give ONE "anchor point" -- pick a pixel
coordinate (x, y) that clearly sits ON the missed content (the
darkest text glyph, the centre of a shape, the middle of a photograph,
etc.). SAM will use this point to figure out the full mask + bbox.

DO report:
- Standalone text labels / paragraphs not in any red rectangle.
- Output / terminal callouts of a flowchart connected by an arrow.
- Sub-elements of a diagram that ended up uncovered.
- A standalone photograph / chart / illustration not in any red
  rectangle.
- Any content a human would consider its own component.

DO NOT report:
- Empty page margins, decorative strips along edges with no content.
- Faint background gradients with no glyphs / strokes.
- Anti-aliasing fringes around an existing red rectangle.
- A point that lands on EMPTY whitespace adjacent to an already-
  extracted component (e.g. do NOT point at the empty space next to a
  photo claiming it's the "other half" of that photo). Only point at
  pixels that visibly have content.
- Frames / outlines with empty interiors.
- Page-edge decorative bars that already touch an existing component.

Note: do NOT make merge decisions in this call. Report fine-grained
missed regions; a later step will decide whether they should be merged
with existing components.

Pick the point CAREFULLY:
- (x, y) coordinates are in ORIGINAL slide pixel space (top-left=0,0).
- Pick a pixel that is INSIDE the missed content, not on its edge.
- For text content, pick a pixel that lands on a visible glyph (not in
  the gap between letters).
- For photos / illustrations, pick a pixel near the visual centre.

Return STRICT JSON only, no prose, no code fences:

{{
  "missed_regions": [
    {{"point": [x, y], "class": "...", "description": "<one sentence>"}}
  ]
}}

If nothing is missed, return ``{{"missed_regions": []}}``.
\end{prompt}

\clearpage\newpage
\begin{prompt}[Prompt 9: Missed Region Validation]
System:
You QA-check ONE proposed 'missed region' from a slide-decomposition pipeline's layout reviewer. The reviewer claimed it found a real component that the pipeline's main detector missed. Your job is to decide whether the proposed bbox actually contains a real, useful slide component -- or whether it's a mistake (an empty patch, anti-aliasing fringe, a sliver of background, an off-by-far rectangle, a duplicate of an already-extracted component, etc.) that should be discarded.

User:
You are given:

- IMAGE 1: the bbox CROP of the proposed missed region (what would
  appear as the new component's image if we accept the proposal).
- IMAGE 2: the full original SLIDE with the proposed bbox outlined in
  red. (Helps you judge what surrounds the bbox and whether the
  rectangle accurately frames a real component.)

Proposal details:
- proposed class:        "{class_name}"
- proposed description:  "{description}"
- proposed bbox xyxy:    {bbox}

Decide: is this proposal a real, useful slide component that the
pipeline should add?

Mark valid = TRUE when:
- IMAGE 1 clearly shows recognisable content matching (or close to) the
  proposed class -- readable text, a visible shape / icon / arrow,
  a photograph, a chart fragment, a labelled callout, etc.
- The bbox in IMAGE 2 reasonably tightly frames the actual content
  (small bbox padding is fine; large empty padding is NOT fine).
- The content is not already captured by an existing red rectangle in
  IMAGE 2 (i.e. it's genuinely a NEW component, not a duplicate).

Mark valid = FALSE when:
- IMAGE 1 is essentially blank, uniform background, or only contains
  anti-aliasing fringe / edge pixels.
- IMAGE 1 contains content but the bbox is far off -- most of the
  meaningful content sits OUTSIDE the proposed rectangle.
- The proposed region is just empty whitespace adjacent to an already
  extracted component (e.g. the proposal claims "the other half of
  the photo" but there are no photo pixels there).
- The content inside is just a 1-3 character fragment of a larger text
  block that's already extracted elsewhere.
- The content is a thin frame / divider line / decorative edge that
  the pipeline standard says should be ignored.

Return STRICT JSON only, no prose, no code fences:
{{"valid": true|false, "reason": "<one short sentence>"}}
\end{prompt}

\clearpage\newpage
\begin{prompt}[Prompt 10: Perceptual Merge Group Proposal]
System:
You review a slide-decomposition pipeline's component layout. The pipeline tends to extract sub-elements of complex structures (flowcharts, architecture diagrams, multi-panel figures) as many separate small bboxes instead of one coherent component. Your sole job in this call is to decide which existing components should be MERGED into a single semantic component.

User:
You are given:

- IMAGE 1: the original slide.
- IMAGE 2: the same slide with a red rectangle drawn over EVERY current
  fine-grained component the pipeline has decided on. This includes both
  components extracted by SAM3 across all iterations AND components
  newly identified as missed in the previous step. Treat all red
  rectangles uniformly -- there is no semantic difference between them at
  this stage.

Here is the structured list of every red-rectangle component (use these
filenames to refer to them in your output):

{components_json}

Allowed taxonomy classes (use ONLY these exact strings for
`merged_class`):
{taxonomy_json}

Identify groups of 2 or more red rectangles that together cover
sub-elements of ONE coherent semantic component (e.g. multiple boxes /
arrows / labels of one flowchart, multiple cells of one architecture
diagram, multiple icons of one icon-collection figure, a flowchart's
internal blocks plus its terminal output callout label, OR two bboxes
that the pipeline produced as fragments of the same underlying piece
of text).

For each group provide:

- "member_filenames": the filenames of the bboxes to merge (must all
  appear in the components list above; can mix any combination of
  iter-prefixed and missed-prefixed filenames).
- "merged_class": one taxonomy class that names the merged component
  (e.g. "flowchart", "diagram", "architecture diagram", "image
  collection", "bullet list", "footer"). MUST be from the taxonomy list
  above.
- "merged_bbox": [x1, y1, x2, y2] -- the union bbox covering all members.
  Use ORIGINAL slide pixel coordinates.
- "reason": one short sentence.

PRIORITY RULE -- fragment-overlap merge (overrides every "do not merge"
rule below):

- If two or more bboxes OVERLAP HEAVILY -- the smaller is mostly
  contained in the larger (>= ~70%
  larger), OR they share most of their area pairwise -- AND they
  textually/visually represent the SAME content (one is a fragment of
  the other; they are two halves of the same text line; the smaller's
  OCR is a substring of the larger's OCR), they are pipeline
  fragmentation artefacts and MUST be merged. Heavy spatial overlap +
  same/substring content is a strong, deterministic signal -- apply this
  rule even if one or more members are normally "complete units"
  (footer, copyright notice, title).
- Examples that MUST merge under this rule:
  - The last bullet of a list extracted both as "bullet list" (a wide
    bbox covering all bullets) AND as a separate "caption/label" or
    "bullet point text" inside that wide bbox -- merge into one bullet
    list.
  - Two overlapping bboxes at the bottom of the slide both reading parts
    of the same copyright string -- merge into one footer.
  - A "page number" bbox sitting inside a "footer" bbox both covering
    the same trailing text -- merge into one footer.
  - A heading split into two adjacent overlapping bboxes -- merge.

DO merge (in addition to the priority rule):
- Multiple boxes/arrows/labels that together form a single flowchart,
  including its terminal output callout.
- Multiple cells/tiles of a single grid figure.
- A title + sub-headings + a body that together form a single labelled
  panel.

DO NOT merge:
- Visually adjacent but semantically independent components that DO
  NOT overlap heavily (two separate text blocks side by side; bullet
  list to the left of a chart on the right; standalone slide title
  above a separate diagram).
- A whole standalone footer / logo / title with an independent
  component on a different part of the slide. (But DO merge two
  overlapping fragments of the SAME footer -- this is the fragment-
  overlap priority rule.)
- A whole component with another whole component just because they
  share a row or column AND there is no spatial overlap between them.

Return STRICT JSON only, no prose, no code fences:
{{
  "merge_groups": [
    {{"member_filenames": [...], "merged_class": "...", "merged_bbox": [x1,y1,x2,y2], "reason": "..."}}
  ]
}}

If nothing should be merged, return ``{{"merge_groups": []}}``.
\end{prompt}

\clearpage\newpage
\begin{prompt}[Prompt 11: Merge Candidate Arbitration]
System:
You decide whether a single candidate slide-component should be merged into an already-formed semantic merge group. The pipeline merged some components into a coherent unit; a non-member candidate now overlaps the group's region. Make a per-candidate yes/no merge decision.

User:
You are given:

- IMAGE 1: the original slide.
- IMAGE 2: the slide with two highlighted regions:
  - GREEN polygon outline = the current merge group's region (members listed below).
  - YELLOW thick rectangle = the CANDIDATE component being considered for inclusion.

Existing merge group (already a coherent unit):
- merged_class: "{merged_class}"
- members:
{merged_members_json}

Candidate:
- filename: {candidate_filename}
- class: {candidate_class}
- bbox (x1,y1,x2,y2): {candidate_bbox}

Decide: should the CANDIDATE be merged INTO this group, becoming part of
the same semantic component?

Set should_merge = TRUE when (any of):
- The candidate is the missing complement of the merged unit (e.g. the
  RHS of an equation whose LHS is in the group; a legend that explains
  chart elements in the group; a caption beneath a figure in the group;
  the terminal callout of a flowchart whose body is in the group).
- The candidate is a fragment of the same content as a member (e.g. a
  heading split into two adjacent overlapping bboxes; two halves of the
  same footer line).
- Removing the candidate would make the merged group semantically
  incomplete.

Set should_merge = FALSE when:
- The candidate is an independently meaningful component (a separate
  text block, a different figure, an unrelated label) that just happens
  to sit inside the merged group's bounding rectangle because the group
  is wide / tall / spans much of the slide.
- The candidate's content is self-contained and does NOT extend or
  complete the merged group.
- They are visually adjacent but semantically independent.

Then ALSO decide ``visually_overlaps``: do the actual pixels of the
candidate share screen space with any member's actual pixels, or do
they only share BBOX space?

Set visually_overlaps = FALSE (just bbox overlap, "L-shape") when:
- The group's bbox is L-shaped or wide/tall and visually wraps around
  the candidate, but the group members' real strokes/glyphs/pixels do
  not touch the candidate's strokes/glyphs/pixels. E.g. an L-shaped
  flowchart whose bbox encloses an unrelated photo in its inner corner;
  a tall bullet list whose bbox extends past the actual text into a
  chart on the right.
- The group's bounding rectangle just spans across the candidate's
  region because of one outlier member, but no member visually overlaps
  the candidate.

Set visually_overlaps = TRUE (real visual overlap) when:
- The candidate sits on top of / inside / behind a member's actual
  content -- a logo or watermark over a photo, an icon embedded inside a
  diagram, a label whose glyphs are drawn directly over a chart's bars.
- Their actual pixels mix in the same screen region.

When should_merge=TRUE, you can set visually_overlaps to whatever is
true; it will be ignored downstream.

Return STRICT JSON only, no prose, no code fences:
{{"should_merge": true|false, "visually_overlaps": true|false, "reason": "<one short sentence>"}}
\end{prompt}

\clearpage\newpage
\begin{prompt}[Prompt 12: Overlap Ownership Arbitration]
System:
You arbitrate ownership of an overlap region between two slide components. Two extracted bboxes overlap on the slide; for each side you decide whether THAT side's actual content extends into the overlap region (true) or whether the bbox just sloppily extends into the other component's territory without owning that region (false). Use the visual evidence on the slide, not just the bbox geometry.

User:
Two extracted slide components have overlapping bboxes. Decide who owns the overlap.

- IMAGE 1: original slide.
- IMAGE 2: same slide with two highlighted bboxes:
  - RED rectangle = component A
  - BLUE rectangle = component B
  - the overlap region is shaded YELLOW.

Component A:
- class: "{class_a}"
- bbox (x1,y1,x2,y2): {bbox_a}

Component B:
- class: "{class_b}"
- bbox (x1,y1,x2,y2): {bbox_b}

Look at IMAGE 1 inside the overlap region (the yellow shaded area in IMAGE 2)
and decide for each side independently:

- ``a_owns_overlap``: TRUE iff component A's actual visual content (its
  text glyphs, graphic strokes, photo pixels, etc.) is genuinely present
  inside the overlap region -- i.e. that region is part of A. FALSE when
  A's bbox just sloppily extends into the overlap area while A's real
  content sits elsewhere (typical for text classes whose bbox is too
  wide; the empty bbox padding overlapping another component does NOT
  count as A owning the region).

- ``b_owns_overlap``: same question for B.

Decision matrix:
- TRUE / TRUE  -> both genuinely overlap (legitimate co-located content). No carve.
- TRUE / FALSE -> A owns the overlap; B's bbox is the suspect side. Pipeline will carve B by A.
- FALSE / TRUE -> B owns the overlap; A's bbox is the suspect side. Pipeline will carve A by B.
- FALSE / FALSE -> ambiguous / unrelated; pipeline will leave both bboxes unchanged.

Common patterns:
- A is a bullet list / body text whose bbox extends past the actual text into a photograph or chart on the right: a_owns=false, b_owns=true -> carve A.
- A is a slide title whose bbox extends into a logo / icon on the right: a_owns=false, b_owns=true.
- A photo and a small caption beneath it that genuinely co-located in the overlap: typically the smaller bbox is fully inside the bigger but each owns its part of the overlap. Use TRUE/TRUE only if both clearly have content there.

Return STRICT JSON only, no prose, no code fences:
{{"a_owns_overlap": true|false, "b_owns_overlap": true|false, "reason": "<one short sentence>"}}
\end{prompt}

\clearpage\newpage
\begin{prompt}[Prompt 13: Segmentation Order Planning]
System:
You plan the SAM3 segmentation order for one slide. The pipeline will call SAM3 once per LAYER in the order you produce. Each layer is a SEPARATE SAM3 call -- its detector sees ONLY the classes in that layer, and SAM3's built-in cross-class dedup only suppresses duplicates within a single call. So if you split classes that describe the same visual content into different layers, both will fire and produce duplicate bboxes. KEEP classes that target the same visual content in the SAME layer.

Each layer after the first runs on the slide with all earlier layers' mask pixels already removed. Order classes so that the ones most prone to absorbing UNRELATED neighbouring pixels are segmented LAST (after their neighbours are gone), and the ones most likely to be ABSORBED BY a wider class go EARLIER (so they are claimed before the wide class fires).

You DO NOT add, remove, or rename classes -- output the exact same set, just partitioned and ordered.

User:
You are given:

- IMAGE: a presentation slide.
- CLASS LIST (every class to segment on this slide):

{class_list_json}

Partition the class list into 2 OR 3 ordered LAYERS (NOT 4 unless the
slide truly has four visually distinct depth tiers -- most slides need
just 2). Layer 1 is segmented first on the original slide. Layer 2
runs on the cleaned image after Layer 1's masks are removed. And so on.

PRIMARY GOAL: keep classes that pick up the SAME visual region in the
SAME layer, so SAM3's internal cross-class dedup can suppress
duplicates. If a class is split off into a later layer where SAM3 only
sees it (no competing class), SAM3 will produce a fresh bbox even when
the same region was already extracted -- this is by far the worst
failure mode of layered seg.

SECONDARY GOAL: prevent mask leakage by ordering. When mode (a) and
mode (b) both apply, pick whichever is the bigger risk for THIS slide:

  (a) A small / on-top / tightly-bordered class is at risk of being
      ABSORBED by a larger class's mask -> put the small one EARLIER.
      Examples: a math symbol on top of a chart, a callout label over
      a photograph, brackets / single characters adjacent to a wide
      illustration.

  (b) A small / well-bordered class's mask is at risk of HALLUCINATING
      adjacent decorative pixels into itself (e.g. a "bullet list"
      mask grabbing the green edge of an area chart that sits beside
      it; a "logo" mask grabbing background gradient pixels) -> put
      the WIDE adjacent class EARLIER (so the adjacent decoration is
      cleared before the small class is segmented).

CRITICAL RULES -- break any of these and the layered pipeline will
produce worse output than no layering at all:

R1. Generic / catch-all classes that have weak visual priors
    ("decorative element", "frame", "whitespace", "shape", "icon
    collection", "mixed content", "composite figure", "graph diagram",
    "diagram", "figure", "illustration", "label", "text", "annotation"
    when used as a generic catch-all) MUST share a layer with at least
    one concrete class. NEVER put any of them alone in their own (late)
    layer -- SAM3 will hallucinate slide-spanning false-positive bboxes
    against the emptied image.

R2. Classes that overlap heavily in their visible region MUST be in
    the same layer. Examples to detect from the IMAGE:
    - Multiple bullet-related classes ("bullet list", "bullet marker",
      "bullet point", "bullet point text") all targeting one bullet
      block -> same layer.
    - "Venn diagram" + "bubble chart" both targeting one set of
      overlapping circles -> same layer.
    - "axis" + "axis label" + "axes" + "axis labels" -> same layer.
    - "arrow" + "arrow annotation" -> same layer.
    - "photograph" + "image" or "figure" if both describe the same
      photo -> same layer.

R3. If after applying R1 and R2 you cannot find a clean visual depth
    split, just return ONE layer with every class. One layer is
    strictly better than a bad multi-layer split.

R4. Each layer should contain at least 2 classes (or be the only
    layer). Single-class layers triggered by R1/R2 should be merged
    with the next layer.

Output schema (strict JSON, no prose, no code fences):

{{
  "layers": [
    {{"layer": 1, "classes": ["...", "..."], "reason": "<one short sentence rooted in what's visible on THIS slide>"}},
    {{"layer": 2, "classes": ["..."], "reason": "..."}}
  ]
}}

If you cannot confidently stratify the slide, return:

{{"layers": [{{"layer": 1, "classes": [<every input class>], "reason": "no clear stratification -- better to let SAM3 dedup across all classes in a single call"}}]}}
\end{prompt}

\clearpage\newpage
\subsection{Baseline Details}
\label{app:decompose-baseline-details}
\subsubsection{SlideCoder}
\label{app:decompose-slidecoder-setup}
We evaluate SlideCoder~\citep{tang2025slidecoderlayoutawareragenhancedhierarchical} as a key baseline by reconstructing PowerPoint presentations (\texttt{.pptx}) from visual references. The process begins by taking the slide images as input to the SlideCoder framework, which utilizes VLM to generate corresponding \texttt{python-pptx} scripts. The generated scripts are then programmatically executed using a dedicated driver to synthesize physical \texttt{.pptx} files. Finally, the reconstructed slides are rendered back into high-resolution images to enable downstream automatic evaluation alongside our proposed method.

\subsubsection{Claude Opus 4.7}
\label{app:decompose-claude}

We evaluate Claude Opus 4.7 as a strong proprietary agentic baseline. Rather
than using the Claude API directly, we use Claude Code to assess the capability
of its built-in coding and execution harness. For each deck, a Claude Code
monitor agent launches a worker agent with the prompt below.
The worker agent then reasons over the slide decomposition task, reconstructs
the slide state, and outputs an editable PPTX artifact.

\begin{prompt}[{Baseline Prompt: Claude Opus 4.7 Slide Decomposition}]
Slide Decomposition Prompt

Reconstruct slide images into editable .pptx files. Do NOT take the shortcut of embedding the source PNG full-bleed as a single picture.

Objective

For each input image, build a .pptx that is visually identical to the source when rendered: same layout, text content, fonts, colors, sizes, positioning, and visual elements. Text must be reproduced as real editable text boxes, not images. Use python-pptx text boxes, shapes, and connectors for typography and vector content. For regions you genuinely cannot re-vectorize, such as photos, intricate bitmap diagrams, logos, or charts, crop just that region with PIL and embed it as a picture in its correct slot.

Inputs and outputs

You will be given:

- An input directory containing one or more slide images, such as PNG or JPEG files.
- An output directory where the corresponding .pptx files must be saved.
- A path to the manifest JSON for progress tracking.

For each input <NAME>.png or <NAME>.jpeg, the corresponding output is <output_dir>/<NAME>.pptx. Create the output directory if it does not exist.

Per-image workflow

1. Read the image with the multimodal Read tool so you can actually see it.

2. Observe the layout: background, title, body text, bullets, tables, shapes, arrows, image regions, font choices, weights, alignments, approximate positions and sizes. Note the image's pixel dimensions, which determine the target slide canvas. Use 9144000 x 6858000 EMU, i.e., 10" x 7.5", for 4:3 sources by default, and use 12192000 x 6858000 EMU for 16:9 sources.

3. Write a namespaced Python script using python-pptx and PIL that reconstructs the slide. Follow these conventions:
   - Use Pt(), Inches(), and Emu() for sizes.
   - Reproduce text verbatim with matching font, size, weight, color, and alignment.
   - Recreate vector shapes, such as rectangles, ovals, arrows, connectors, and lines, with correct fills, lines, and positions.
   - For complex raster regions, crop with PIL and embed them as pictures at the correct position and size.

4. Run the script with python3.

5. Verify the output: the file must exist, have size greater than 1024 bytes, and its first four bytes must be b'PK\x03\x04', since every valid .pptx is a zip file. If invalid, retry up to two more times with a simpler approach, using fewer shapes, fewer embedded crops, or a simpler layout.

6. Update the manifest after each image with the locked snippet below.

Manifest update, parallel-safe

The manifest is shared across concurrent agents. Always update it under an exclusive file lock. Substitute <DECK_ID>, <NAME>, <STATUS>, <OUTPUT_PATH>, <ERROR_OR_EMPTY>, and <MANIFEST_PATH>. The value of <STATUS> must be success or failed.

python3 - <<'PY'
import fcntl, json, os
from pathlib import Path

M = Path("<MANIFEST_PATH>")
LOCK = M.with_suffix(".json.lock")
LOCK.touch()

with open(LOCK, "r+") as lf:
    fcntl.flock(lf, fcntl.LOCK_EX)
    try:
        m = json.loads(M.read_text()) if M.exists() else {"decks": {}}
        d = m["decks"].setdefault("<DECK_ID>", {})
        d.setdefault("slides", {})["<NAME>"] = {
            "status": "<STATUS>",
            "output": "<OUTPUT_PATH>",
            "error":  "<ERROR_OR_EMPTY>",
        }
        tmp = M.with_suffix(".json.tmp." + str(os.getpid()))
        tmp.write_text(json.dumps(m, indent=2, sort_keys=True))
        os.replace(tmp, M)
    finally:
        fcntl.flock(lf, fcntl.LOCK_UN)
PY

When all images are processed

- Use the same locked block to set the deck-level entry's status to success if all images are successful, or partial if some failed.
- Set finished_at to the current ISO timestamp.
- Print exactly one line as the very last line of your reply: a compact JSON summary, with no preamble, postscript, explanations, or markdown blocks.

{"deck": "<DECK_ID>", "ok": N, "fail": M, "outputs": ["..."]}

Constraints

- Do not commit anything to git. Do not push.
- Write only to /tmp, the assigned output directory, and the manifest.
- Stay focused on the assigned input set. Do not touch other agents' work.
- python-pptx and Pillow are already installed. Do not pip install anything.
\end{prompt}

\clearpage\newpage
\subsection{Prompt Used for VLM-judge in Evaluation Paradigm}
\label{app:judge-prompt-decompose}
This section provides the VLM-judge (Gemini-3.5-flash) prompts used to evaluate component validity and holistic state quality in slide decomposition.

\begin{prompt}[Component Validity Prompt]
System:
You are an evaluation judge for a slide-decomposition system. You receive one cropped image at a time and a label the system assigned to it. Rate strictly on two axes from 1 to 5 and return JSON only.

User:
The cropped image is the system's prediction of one slide component.
The system labelled it: "{label}".

Rate:
  cohesion (1-5): Is the crop a single coherent semantic unit? 5 = clean single unit (e.g. one chart, one title, one merged flowchart). 3 = mostly fine but with stray fragments. 1 = multiple unrelated things or noise.
  label_fit (1-5): Does the label describe the dominant content? 5 = perfect. 3 = vague but not wrong. 1 = wrong.

Return JSON: {{"cohesion": int, "label_fit": int, "reason": short string}}.
\end{prompt}
\begin{prompt}[Holistic State Quality Prompt]
System:
You are an evaluation judge for slide-decomposition. You will see two images:
1. The ORIGINAL slide annotated with numbered, colored bounding boxes (e.g. #1, #2).
2. The RECONSTRUCTION built by pasting each predicted component crop onto a blank canvas at its predicted bbox, annotated with matching numbered, colored bounding boxes.
Use these visual annotations to judge the decomposition's coverage, granularity, and redundancy. CRITICAL RULES FOR OVERLAPS AND FRAGMENTATION:
- Be TOLERANT of fragmentation and overlaps if they serve the purpose of "editability" in presentation software. For example, separating text from its background container shape, or decomposing complex graphics into layered nested shapes, is highly desirable for editing and should NOT be penalized.
- ONLY penalize UNNECESSARY fragmentation (e.g., a single continuous sentence broken into individual words, or a coherent flowchart shattered into meaningless unrecognizable bits) and UNNECESSARY redundancy (e.g., completely duplicate cloned boxes capturing the exact same content multiple times, or bad layouts that visually occlude distinct text).
In your JSON response, always refer to specific components by their exact IDs (e.g., '#3 overlaps with #5', or '#1 is a merged chart title') in the 'reason' field. Return JSON only.

User:
Image 1 is the original slide with numbered bounding boxes. Image 2 is the reconstruction canvas with matching numbered bounding boxes.
The predicted components list is:
{components}

Rate each axis 1-5:
  coverage (1-5): Are all meaningful slide elements present in the reconstruction? 5 = nothing important missing. 1 = many elements absent.
  granularity (1-5): Are the chosen units the right grain? 5 = grain is perfect. 1 = unnecessarily overly fragmented or wrongly fused. NOTE: Be forgiving of fragmentation if it separates text from shapes or creates editable nested graphical layers. Only penalize senseless shattering (e.g. single sentences broken into words, or unified charts broken into 50 tiny uneditable lines).
  non_redundancy (1-5): Does the system avoid duplicate components and erroneous overlaps? 5 = no erroneous redundancy. 1 = many duplicates or severe layout occlusions. NOTE: Do NOT penalize overlaps that represent logical layering (e.g. text layered over a background box, or nested graphics). Only penalize actual duplicate detections of the identical visual content or destructive occlusions.

Return JSON: {{"coverage": int, "granularity": int, "non_redundancy": int, "reason": short string (<80 words, must cite specific visual component IDs like #1, #2 to justify your ratings)}}.
\end{prompt}

\clearpage\newpage
\subsection{Slide Decomposition Qualitative Results}
\label{app:decompose_qual}

\begin{figure*}[t]
\centering
\includegraphics[width=\linewidth]{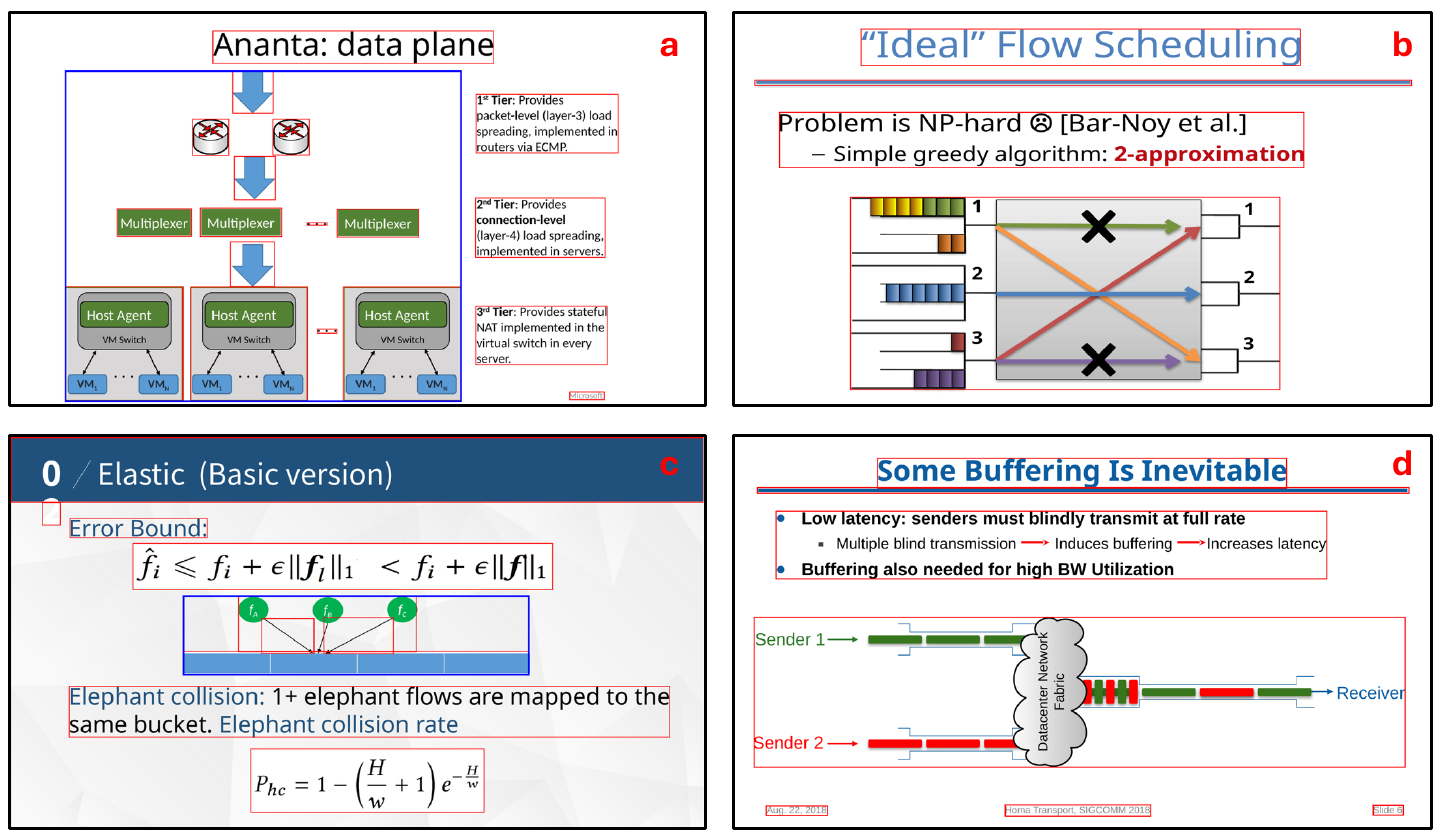}
\caption{
Qualitative results for slide decomposition.
Panels (a)--(d) show representative academic slides with diverse layout structures, including hierarchical diagrams, flow-scheduling figures, equations, and network illustrations.
Red boxes denote the initial component proposals grounded by SAM3.
In panels (a) and (c), blue boxes highlight perception-aligned editable components produced by our decomposition pipeline after consolidating compatible local proposals.
Across these examples, the raw proposals provide dense local grounding but may split a semantic slide object into fragments or include overlapping regions.
Our pipeline consolidates these proposals into higher-level units that better match human perception, such as complete diagrams, equations, titles, and text blocks.
}
\label{fig:decompose_qual}
\end{figure*}

Figure~\ref{fig:decompose_qual} visualizes representative slide decomposition results.
The red boxes show the initial SAM3-grounded component proposals, while the blue boxes show the final perception-aligned editable components produced by our pipeline.
These examples illustrate how dense local grounding is converted into higher-level slide units that better match human perception and support native editing.

\clearpage\newpage

\section{Style Adaptation Pipeline Details}
\label{sec:styleadpapp}

\subsection{Detailed evaluation results of style adaptation}
\label{sec:styleadpapp1}

\begin{figure}[htbp]
\centering
\includegraphics[width=\linewidth]{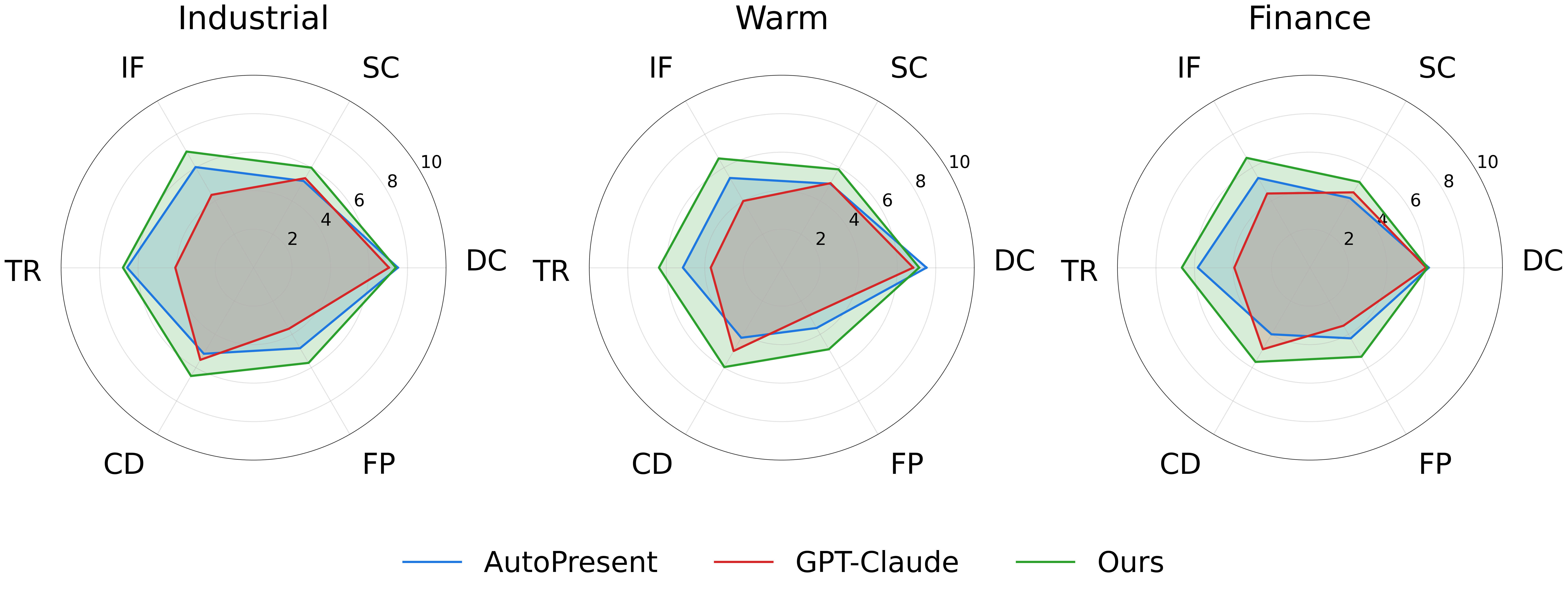}
\caption{Six dimension VLM as a judge results for style adaptation. The abbreviations IF, SC, TR, DC, CD, and FP correspond to Information Fidelity, Slide level Coherence, Text Readability, Deck level Coherence, Content and Decoration Balance, and Fitness for Purpose.}
\label{radar}
\end{figure}

Figure~\ref{radar} summarizes the six dimension VLM as a judge scores across the three styles. \methodname{} attains the largest radar area everywhere, with the clearest gains on the dimensions that demand preserving source information while changing appearance, namely information fidelity, text readability, content and decoration balance, and fitness for purpose. This supports treating restyling as structured artifact transformation rather than full slide image generation. The GPT and Claude baseline produces styled outputs but loses fidelity and readability because it regenerates each slide as an image before recovering structure, while AutoPresent reaches competitive deck level coherence yet its weaker slide level coherence shows that global consistency alone is insufficient. More detailed results are available in section More cases available in section \ref{sec:styleadpapp1}.

\subsection{Another Example case of style adaptation}
\label{sec:styleadpapp2}
\begin{figure*}[!t]
\centering
\includegraphics[width=\linewidth]{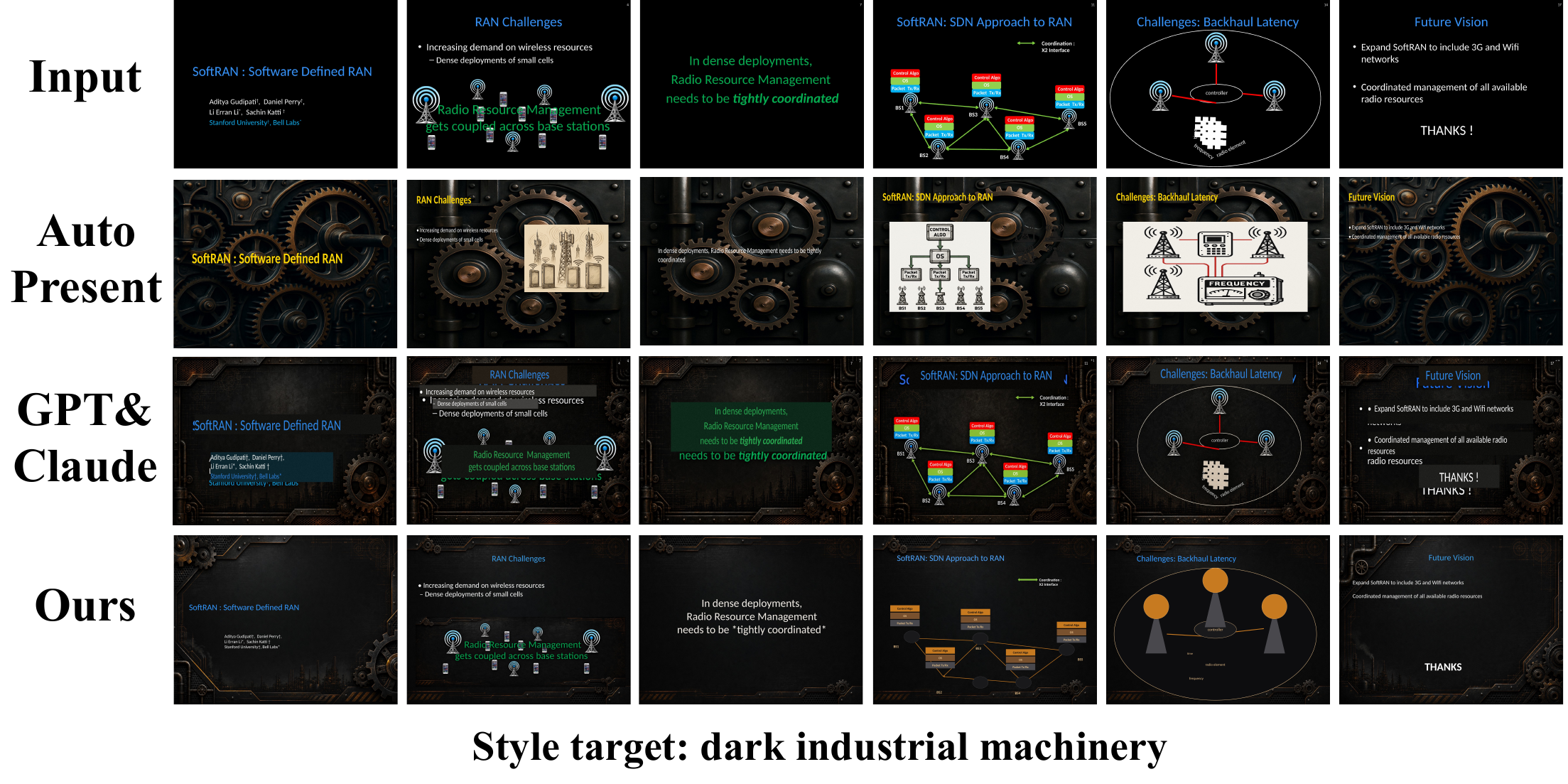}
\caption{Industrial Style Case study: SlideForge outperforms the baselines.}
\label{caseshowapd}
\end{figure*}

Figure~\ref{caseshow} shows another successful case study on six slides under a dark industrial machinery style target. \methodname{} repaints the deck with a dark metallic theme and warm orange accents while keeping every title, bullet, and diagram legible and faithful to the input, and it restyles the SoftRAN control flow and the backhaul topology as native shapes rather than flattening them into pictures. AutoPresent stamps a heavy gear photograph onto every slide, but the busy background overwhelms the text and its diagrams collapse into generic clip art that no longer conveys the original structure. The GPT and Claude pipeline reaches a plausible industrial look yet duplicates and overflows text across several slides, visible in the repeated Future Vision heading and the doubled THANKS line. The comparison shows that \methodname{} adapts the visual theme while preserving the readability, diagram semantics, and layout that the baselines lose.

\subsection{Limitations of style adaptation}
\label{sec:styleadpapp3}

\begin{figure*}[!t]
\centering
\includegraphics[width=\linewidth]{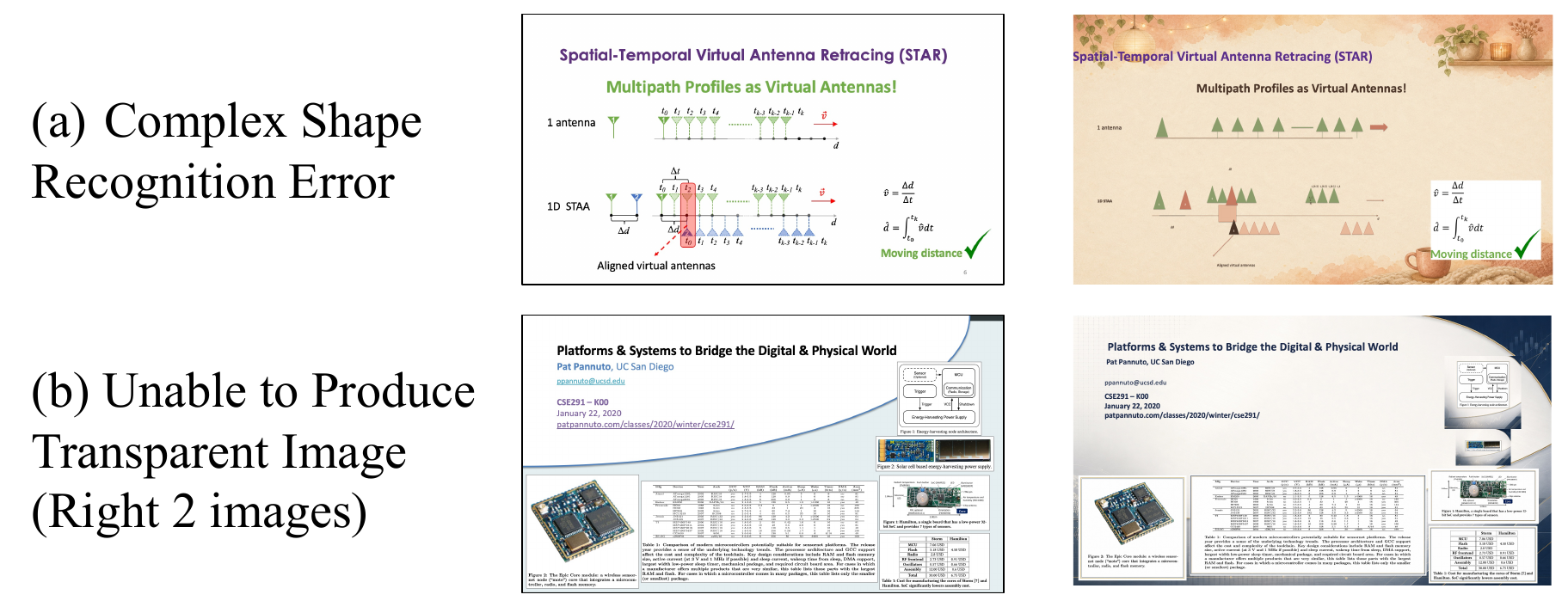}
\caption{Failure Cases.}
\label{fig:badcase}
\end{figure*}

Figure~\ref{fig:badcase} illustrates these two failure modes. In the complex shape recognition example, the $\tau=4$ shape composition modality misreads the densely packed antenna icons of the original STAR slide and rebuilds them as generic triangular marks, so the output keeps the layout but corrupts the meaning of the diagram. In the transparency example, the $\tau=5$ raster path cannot obtain transparent assets from GPT-Image-2, so the embedded figures and the sensor board photo retain an opaque rectangular fill that sits awkwardly on the restyled background. Both cases stem from the limited capacity of the base model rather than from the harness, and stronger control over the VLMs would mitigate them.

\subsection{Prompt Used in Style Adaptation Pipeline}
\label{app:style-prompts}
Our style adaptation pipeline runs in six consecutive steps, step 0
through step 5. Step 0 only organises the upstream decomposition
output into a uniform per-page layout and issues no model call. The
five remaining steps drive a mixture of Claude Opus 4.7 prompts, used
for analysis, decomposition planning, and verification, and
GPT-image-2 prompts, used for styled background generation and
class 5 foreground restyling. Where a single step uses both models,
the Claude prompt and the GPT-image-2 prompt are shown in separate
blocks below.

\paragraph{Step 0. Front-bg Preparation.}
We materialise a uniform per-page directory containing the original
slide, the cleaned background, the segmented foreground components,
and a metadata file with per-component bounding boxes. No language
or image model is queried at this step, so no prompt is shown.

\paragraph{Step 1. Background Style Generation and Palette.}
A short Claude prompt first picks one interior slide as the content
exemplar so that GPT-image-2 sees a representative middle page when
it generates the three role-specific styled backgrounds. Three
GPT-image-2 calls then produce the opening, content, and closing
backgrounds. Finally a Claude prompt extracts a unified theme palette
from the three styled backgrounds, which step 4 will later use as the
allowed color universe.

\begin{prompt}[Claude Opus 4.7 -- Content-Page Picker]
You are choosing which MIDDLE page of a slide deck to use as the "content" exemplar that will be sent to a style-transfer image model alongside the opening and closing pages.

For each candidate above you saw the ORIGINAL slide image and the CLEANED background after the upstream segmenter removed foreground.

Pick the candidate whose CLEANED background best satisfies BOTH
  a. cleanest reference with minimal residual foreground artifacts,
  b. most representative of a typical content slide for THIS deck.

Reply with STRICT JSON only.
{"pick_index": <1-N>, "reason": "<one short sentence>"}
\end{prompt}

\begin{prompt}[GPT-image-2 -- Opening Background]
You are given 2 input images for the OPENING / TITLE page of a multi-page PowerPoint deck.
  Image 1, the CLEANED background of this page.
  Image 2, the ORIGINAL page. Use it ONLY to see where foreground elements such as title, authors, and logos will sit, so your generated decorations do NOT intrude into those bbox regions.

Output requirements.
- Output EXACTLY ONE full-canvas styled background that fills the ENTIRE output image, edge to edge.
- Do NOT split the output into thumbnails, panels, grids, stripes, side-by-side variations, or composites. The output IS the slide canvas, one slide, one output image.
- Decorations belong along the edges, corners, and margins. Leave the large central region visually quiet so the title and authors can be placed on top later.
- This is the opening / title page, so decoration can be on the more elaborate hero end.
- Image 1 may still contain residual logos or text that the upstream segmenter failed to remove. IGNORE them and do NOT preserve them in your output.
\end{prompt}

The CONTENT and CLOSING role prompts share the same scaffold as the
opening prompt, with one extra input image set to the already
generated opening so the deck inherits a consistent palette and motif
vocabulary across slides while differing in composition. The CONTENT
prompt additionally constrains decoration to the top and bottom edges
only and demands a quiet central band, since interior slides routinely
host bullets, diagrams, and charts that extend toward the side
margins.

\begin{prompt}[Claude Opus 4.7 -- Palette Extractor]
You are a senior visual designer analyzing the visual style of a slide deck. You have just been shown THREE slide background images from the SAME deck, ordered as opening, content, closing.

Treat the three images as ONE coherent visual system and derive a unified design palette that describes the deck as a whole rather than each image individually. Produce.

1. THEME COLOR. The single most representative dominant color. Provide an uppercase hex string and a short human-readable name.
2. THEME PALETTE. An ordered list of 4 to 6 colors from MOST prominent to LEAST prominent. Give hex, name, and a one-phrase role for each entry, such as "background", "primary accent", or "supporting tone".
3. FONT PALETTE. An ordered list of recommended font colors to use ON TOP of the theme. Sort STRICTLY from highest contrast to lowest contrast. For each entry give hex, name, an approximate WCAG contrast ratio, and a one-phrase recommended use.

Respond with STRICT JSON only. No markdown fences and no commentary.
\end{prompt}

\paragraph{Step 2. Modality-Typed Element Classification.}
For each foreground component prepared in step 0, a Claude prompt
assigns one of six modality classes that decides which reconstruction
primitive is used downstream. The classifier is conservative on
semantically meaningful elements and never sends them to the discard
class.

\begin{prompt}[Claude Opus 4.7 -- Component Classifier]
You are analyzing ONE foreground element extracted from a PowerPoint
slide.

The FIRST image is the ORIGINAL FULL SLIDE. Use it as context for the element's role within the slide.
The SECOND image is the FOREGROUND ELEMENT to classify, isolated from the slide on a transparent background.

The filename hint produced by an upstream detector is "<<FILENAME_HINT>>". The hint is informative but NOT guaranteed.

Choose ONE class.
  0 decorative content with no semantic meaning, drop.
  1 pure text or formula without a background fill.
  2 text or formula on a colored or shaped background.
  3 structured text that must remain editable, typically a table.
  4 shape composition approximable by primitive PPTX shapes.
  5 raster-only content such as photo, complex logo, screenshot, or
    data chart.

Be conservative. Elements that may carry semantic content, such as logos, charts, labels, arrows, annotations, or page numbers, must NOT be assigned to class 0.

Respond with STRICT JSON only.
{"class": <0-5>, "reason": "<one short sentence>"}
\end{prompt}

\paragraph{Step 3. Class-Conditional Reconstruction.}
Every class above 0 has its own reconstruction driver. Classes 1, 2,
3, and 4 are reconstructed natively as editable PPTX shapes following
a shared render-and-verify loop where a VLM proposal is rendered
headlessly and the rendering is compared against the original bbox
crop. Class 5 is restyled through a GPT-image-2 image edit guided by
a Claude-written style prompt. We list one representative prompt per
class plus the shared verifier.

\begin{prompt}[Claude Opus 4.7 -- Class 1 Text Extractor]
You are looking at one foreground element extracted from a PowerPoint slide.

Image 1, the ORIGINAL FULL SLIDE for context.
Image 2, the FOREGROUND ELEMENT to encode as an EDITABLE TEXT BOX.

This element was classified as class 1, pure text or formula WITHOUT any background colour or shape. Reproduce it faithfully as a single textbox that fits EXACTLY inside its bounding box. Decide.

- the verbatim text content. Preserve casing, punctuation, spaces,
  and any line breaks. Use \\n to encode line breaks.
- the TOTAL number of visible text lines.
- the font family.
- the font size in points.
- the alignment.
- the font color drawn from the deck FONT palette.

Respond with STRICT JSON only.
\end{prompt}

\begin{prompt}[Claude Opus 4.7 -- Render-and-Verify Loop]
You are verifying a PPTX textbox reconstruction.

Image A, the ORIGINAL foreground element as ground truth.
Image B, the CURRENT textbox rendered from the PPTX, cropped to the same region as the current textbox bbox.

Constraints in priority order.
  1. HARD. The rendered text must NOT extend beyond the textbox bbox.
  2. HARD. The rendered font size must NOT go below <<MIN_PT>> pt.
  3. SOFT. Aim to match the original line count <<ORIG_NUM_LINES>>.

If Image B violates any HARD constraint, propose a minimal change. Either extend the bbox WIDTH by no more than <<MAX_EXTEND_PX>> pixels inside the slide canvas, or shrink the font by 1 pt, or both. If Image B already satisfies all HARD constraints and the line count is close enough, accept.

Respond with STRICT JSON only.
\end{prompt}

\begin{prompt}[Claude Opus 4.7 -- Class 3 Table Structure Pass]
You are looking at the FOREGROUND TABLE element to encode as a NATIVE, EDITABLE PPTX TABLE.

THIS PASS returns ONLY the table STRUCTURE plus a small global STYLE ANCHOR. The cell content is requested in a separate batched call so single-call output never truncates. Provide.

- grid size n_rows and n_cols.
- header_row and first_col_header boolean flags.
- col_widths_frac and row_heights_frac, fractional sequences that sum to 1.
- a style anchor with default cell fill hex, header fill hex,
  border hex, and body text color hex drawn from the deck palettes.

Respond with STRICT JSON only.
\end{prompt}

\begin{prompt}[Claude Opus 4.7{,} Class 4 Shape Composition]
You are looking at a FOREGROUND ELEMENT to approximate as a composition of NATIVE, EDITABLE PPTX SHAPES.

Decompose into the SMALLEST set of primitive PPTX shapes that recognizably reproduces the element, at most <<MAX_SHAPES>> shapes. Use a fractional coordinate system rel_xyxy in [0, 1] inside the element bounding box. The build step hard-clamps every output shape inside the bbox, so any positional drift is automatically corrected and never produces overflow.

Respond with STRICT JSON only.
\end{prompt}

\begin{prompt}[Claude Opus 4.7 -- Class 5 Image-Edit Prompt Generator]
You are preparing a single prompt that will be sent to GPT-image-2 to perform STYLE TRANSFER on a class 5 foreground element such as a logo, photograph, text-and-image composite, or data chart.

Inputs you saw are the original full slide, the foreground element itself, and the new styled background. Produce a self-contained instruction that asks GPT-image-2 to restyle the element so it harmonises with the new background while preserving any identity or data-critical content. Logo wordmarks must remain readable, chart axis labels and tick values must remain identical, and screenshot text must remain identical. Do NOT add new icons, captions, or decorative elements.

Respond with STRICT JSON only, carrying a single field "image_prompt" whose value is the natural language prompt for GPT-image-2.
\end{prompt}

\begin{prompt}[GPT-image-2 -- Class 5 Foreground Restyle]
<<IMAGE_PROMPT EMITTED BY CLAUDE IN THE PRECEDING STEP>>

Example payload for a warm-cozy chart, "Restyle this matplotlib line chart so it harmonises with the cream and terracotta parchment ground. Keep all axis ticks, axis labels, legend entries, line colors, and numerical values identical to the input chart. Replace only the chart background and axis line colors with muted amber. Do NOT redraw the data, do NOT add new labels, and do NOT introduce decorative props."
\end{prompt}

\paragraph{Step 4. Palette-Constrained Color Restyling.}
Each non-raster reconstruction from step 3 is recoloured so it
harmonises with the new styled background. Geometry, font metrics,
and table dimensions from step 3 remain untouched. Claude builds a
functional color map under the rule that identical source hex values
must map to identical target hex values, so semantic equivalence
classes such as header vs body cells survive the recolor.

\begin{prompt}[Claude Opus 4.7 -- Table Recolor]
You are restyling a PPTX TABLE so it harmonises with a NEW slide background.

Given.
  Image A, the new slide background.
  Image B, the table BEFORE restyling.
  THEME palette, the allowed cell background colours.
  FONT palette, the allowed text colours ranked by contrast.
  Distinct current cell-fill hex values with counts.
  Distinct current cell-border hex values with counts.

Map EACH DISTINCT current cell-fill colour to a NEW hex from the THEME palette, and EACH DISTINCT cell-border colour to a NEW hex from the THEME palette.

Hard rules.
- Identical source hex always maps to the same target hex.
- Pick fills with MODERATE contrast against the new background.
- Header rows must remain visually distinct from body rows.
- Text colors low in contrast against the new cell fill must be
  swapped to the highest-contrast entry from the FONT palette.

Respond with STRICT JSON only.
\end{prompt}

\begin{prompt}[Claude Opus 4.7{,} Shape Composition Recolor]
You are restyling a PPTX SHAPE COMPOSITION so it harmonises with a NEW slide background.

Given the new background, the shape composition, the THEME and FONT palettes, the distinct shape-fill hex values, and the distinct shape-line hex values, map each DISTINCT fill colour and each DISTINCT line colour to a new hex from the THEME palette.

Hard rules.
- Moderate contrast against the new background.
- Identical source hex always maps to identical target hex.
- Any low-contrast text inside the shape must be swapped to the
  highest-contrast entry from the FONT palette.

Respond with STRICT JSON only.
\end{prompt}

\paragraph{Step 5. Final Assembly and Self-Critic.}
For each page the role-specific styled background is placed as the
slide canvas, then every reconstructed component is restored on top,
deep-copied from its step 3 or step 4 PPTX and anchored to the
decomposition-time bounding box. A conservative Claude self-critic
then reviews the assembled deck. It is restricted to two safe
operations, namely block-level z-order changes and text-scale
hierarchy adjustments, and it never adds, removes, or moves slide
content.

\begin{prompt}[Claude Opus 4.7 -- Z-Order Self-Critic]
You are reviewing the Z-ORDER of foreground shape blocks on ONE slide of a restyled PowerPoint deck.

Image A, the ORIGINAL slide BEFORE style transfer. Use it ONLY as ground truth for WHICH element should be ABOVE which. Image B, the CURRENT restyled slide as rendered.

The current slide lists foreground shape blocks from BACK, drawn first, to FRONT, drawn last. Each block is one logical component such as one text block, one image, or one shape group.
<<BLOCK_LIST>>

Only propose changes if the rendering shows a CLEAR occlusion or overlap that contradicts Image A. Never add, remove, or relocate content. Only suggest pairwise z-order swaps or batch moves to back or front.

Respond with STRICT JSON only.
\end{prompt}

\clearpage\newpage
\subsection{Prompt Used for VLM-judge in Evaluation Paradigm}
\label{app:style-judge-prompt}
Our restyling evaluation drives a Claude Sonnet 4.6 judge once per deck. Every rendered slide of the output deck is shown to the judge in slide order, together with the original natural-language style prompt that the deck was meant to inhabit. The judge then scores six independent aspects on a 1 to 10 scale and is required to cite a specific visual element for each score. The exact system-level prompt is reproduced below. The placeholder \texttt{\{\{STYLE\_PROMPT\}\}} is substituted with the per-deck target style description before the prompt is sent.

\begin{prompt}[Claude Sonnet 4.6 -- Style-Adaptation Judge]
You are an expert design critic evaluating style-transferred presentation slides. You will be shown ONE deck of slides produced by a style-transfer approach. The deck was generated to match this TARGET STYLE.

  TARGET STYLE PROMPT. """{{STYLE_PROMPT}}"""

Evaluate the deck on the SIX independent dimensions below.

IMPORTANT RULES.
- Score each dimension INDEPENDENTLY. A deck may score high on
  some and low on others.
- For each score, cite at least ONE specific visual element you
  observed, such as "slide 3 title font" or "chart axis labels
  on slide 5". Refer to slides by the filename label shown next
  to each image, such as "slide_002.png".
- Do not let one dimension contaminate another. A deck with strong
  cross-slide consistency can still fail information fidelity, and
  vice versa.
- Judge purely on what you see. Do not assume anything about which
  method produced the deck.

DIMENSIONS, score each 1 to 10.

1. DECK-LEVEL COHERENCE, cross-slide consistency.
   Looking ACROSS all slides in the deck, do they share one consistent visual identity, namely the same palette, typography, motif, framing, and style treatment. Penalize slides that look like they belong to different decks, abrupt shifts in treatment between slides, or any slide left visually out of step with the rest. This measures slide-to-slide consistency, NOT how well any single slide matches the theme.

2. SLIDE-LEVEL COHERENCE, element-to-theme alignment within a slide. 
    Looking WITHIN each slide, does EVERY element, background, icons, charts, diagrams, text styling, and decorative shapes, conform to the TARGET STYLE PROMPT above. Penalize any element that breaks the theme, such as a default chart pasted onto a themed slide, generic clipart inconsistent with the target aesthetic, raw original assets that the style transfer never reached, or decorative elements that contradict the prompt. This measures element-to-theme fidelity within each slide, NOT consistency between slides. Average your per-slide judgments.

3. INFORMATION FIDELITY.
   Are the original information-bearing elements, including chart data, axis labels, diagram structure, and technical figures, preserved accurately. Penalize hallucinated text, garbled labels, redrawn charts whose numbers no longer match, or diagrams that lost structural meaning.

4. TEXT READABILITY.
   Can all body text, labels, captions, and annotations be read effortlessly. Penalize low contrast, decorative fonts over busy backgrounds, text overlapping imagery, or labels too small for their container.

5. CONTENT-DECORATION BALANCE.
   Does decoration support content, or compete with it. Penalize both extremes, decoration so dominant it crowds the message, AND decoration so sparse the deck feels like a generic template with a sticker.

6. FITNESS FOR PURPOSE, academic and technical talk.
   Would this deck function in its likely use case, a research talk or technical presentation where the audience needs to extract precise information.

OUTPUT FORMAT. Strict JSON only, no prose outside the JSON block.

{
  "deck": {
    "deck_level_coherence":       {"score": <1-10>, "evidence": "<specific observation>"},
    "slide_level_coherence":      {"score": <1-10>, "evidence": "<specific observation>"},
    "information_fidelity":       {"score": <1-10>, "evidence": "<specific observation>"},
    "text_readability":           {"score": <1-10>, "evidence": "<specific observation>"},
    "content_decoration_balance": {"score": <1-10>, "evidence": "<specific observation>"},
    "fitness_for_purpose":        {"score": <1-10>, "evidence": "<specific observation>"}
  },
  "summary": "<2-3 sentences describing the deck's overall strengths and weaknesses against the target style>"
}
\end{prompt}

\clearpage\newpage
\section{Dataset Documentation and Ethical Considerations}
\label{app:dataset_ethics}

\subsection{Data Scope and Coverage}
 We document that our artifacts consist of academic presentation slides, with SAM3 training data collected from AI conference presentations and the out-of-domain test set collected from system-conference presentations. The slides are primarily English and contain typical slide-level language, including titles, bullet points, captions, chart labels, legends, equations, tables, and short explanatory text. The dataset is not intended to represent demographic groups, and we do not collect or annotate demographic attributes. Personal names, when present, correspond to public academic attribution or cited researchers. These characteristics define the intended scope and limitations of the dataset.

\subsection{Privacy and Identifiability}
The collected academic slides may include public attribution information, such as author names, presenter names, affiliations, and cited researchers. We retain such information only when it is part of publicly available academic material and do not use it for person-level identification or profiling. We reviewed the slides for private identifiers, uniquely identifying personal details, and offensive content. Slides containing private contact information, sensitive identifiers, or offensive content were removed or anonymized. Our released annotations focus on slide structure and visual components, including bounding boxes and semantic labels, rather than personal information.

\subsection{Content Safety}
We did not conduct a systematic screening for offensive content; because the corpus is drawn exclusively from peer-reviewed academic conference material (both the AI/ML training set and the systems/networking OOD test set), we consider the likelihood of offensive content to be negligible.

\section{LLM Use}
\label{app:llm-use}
We used LLMs for writing polishing and implementation support. LLMs were also used within the proposed pipeline for prompt-based component labeling and data annotation. All LLM-generated or LLM-assisted outputs were manually reviewed by the authors before use.

\section{Human Evaluation Interface and Instructions}
\label{app:human-eval-instructions}

Before beginning the evaluation, participants were asked to provide basic
demographic and background information, including age, gender, highest
education level, field of study or work, English reading proficiency,
frequency of making presentation slides, and familiarity with AI-based
image or slide generation tools. Name was used to save and resume study
progress, while email was optional. Participants were also asked to consent
to the use of their anonymized responses and demographic information for
academic research.

The following instruction text was presented verbatim to participants:

\begin{prompt}[Slide Quality --- Human Evaluation Study]
Thank you for participating! You will look at presentation slides and rate
their quality. The study has three parts: Part 1 --- rate restyled slide
decks (20 sets, the first 10 are required). Part 2 --- rate how well slides
were split into editable components (20 slides, first 10 required). Part 3
--- judge single slides: human-made or AI-rebuilt? (20 slides, first 10
required). The required items take about 45 minutes; anything beyond them is
an optional bonus --- every extra answer helps us. Your progress is saved
after every question: you can close the page anytime and continue later by
entering the same name. There are no right or wrong answers --- we want your
honest visual judgment. Please work on a reasonably large screen (laptop or
monitor, not a phone). You can click any image to view it full-screen.
\end{prompt}

\label{sec:appendix}

\end{document}